\documentclass[letterpaper]{article} 
\usepackage{aaai25} 
\usepackage{times}  
\usepackage{helvet} 
\usepackage{courier} 
\usepackage[hyphens]{url}  
\usepackage[colorlinks=true, 
            linkcolor = blue,
            urlcolor  = blue,
            citecolor = blue,
            anchorcolor = blue
            ]{hyperref} 
\usepackage{graphicx} 
\usepackage{natbib} 

\usepackage[utf8]{inputenc}
\usepackage[T1]{fontenc}

\usepackage{caption} 
\usepackage{algorithm}
\usepackage{algorithmic}
\usepackage{booktabs}
\usepackage{placeins}
\usepackage{listings}
\usepackage{appendix}

\usepackage{newfloat}
\usepackage{listings}
\DeclareCaptionStyle{ruled}{labelfont=normalfont,labelsep=colon,strut=off} 
\floatstyle{ruled}
\newfloat{listing}{tb}{lst}{}
\floatname{listing}{Listing}
\newcommand\blfootnote[1]{%
  \begingroup
  \renewcommand\thefootnote{}\footnote{#1}%
  \addtocounter{footnote}{-1}%
  \endgroup
}

\usepackage{chngcntr}

\usepackage{pythonhighlight}
\usepackage{xcolor} 
\usepackage{amsmath}
\usepackage{fontawesome}
\nocopyright

\title{Toxicity of the Commons: Curating Open-Source Pre-Training Data}
\author {
    Catherine Arnett\textsuperscript{*\rm 1,2},
    Eliot Jones\textsuperscript{*$\dagger$\rm 3},
    Ivan P. Yamshchikov\textsuperscript{\rm 1,4},
    Pierre-Carl Langlais\textsuperscript{\rm 1}
}
\affiliations {
    \textsuperscript{\rm 1}PleIAs, Paris, France\\
    \textsuperscript{\rm 2}Department of Linguistics, University of California San Diego
    \\
    \textsuperscript{\rm 3}Gray Swan AI 
    \\
    \textsuperscript{\rm 4}Center for Artificial Intelligence, Technical University of Applied Sciences W\"urzburg-Schweinfurt
    \\
    \texttt{\hphantom{ppppppp}catherine@pleias.fr} \hphantom{pppppp} \texttt{eliot@grayswan.ai\hphantom{pppppppppp}} \\\texttt{ivan.yamshchikov@thws.de} \hphantom{pppppp} \texttt{pierre-carl@pleias.fr}\hphantom{ppppp} \\
    \bigskip
    \begin{center}
\textcolor{red}{\textbf{CONTENT WARNING}: 
This paper discusses topics that may be upsetting.}
\end{center}
}

\begin{document}
\maketitle

\begin{abstract}
Open-source\blfootnote{*equal contribution} large language models\blfootnote{$\dagger$ Work conducted while an intern at PleIAs.} are becoming increasingly available and popular among researchers and practitioners. While significant progress has been made on open-weight models, open training data is a practice yet to be adopted by the leading open-weight models creators. At the same time, there researchers are working to make language models safer. We propose a data curation pipeline to reduce harmful outputs by models trained on public domain data. There are unique challenges to working with public domain data, as these sources differ from web text in both form and content. 
Many sources are historical documents and are the result of Optical Character Recognition (OCR). Consequently, current state-of-the-art approaches to toxicity filtering are often infeasible or inappropriate for open data models. 

In this paper, we introduce a new fully open-source pipeline for open-data toxicity filtering. Our contributions are threefold. We create a custom training dataset, \textbf{ToxicCommons}, which is composed of texts which have been classified across five different dimensions (racial/origin-based, gender/sex-based, religious, ability-based discrimination, and violence). We use this dataset to train a custom classifier, \textbf{Celadon}, that can be used to detect toxic content in open data more efficiently at a larger scale. Finally, we describe the balanced approach to content filtration that optimizes safety filtering with respect to the filtered data available for training. 
    
\end{abstract}

\begin{center}
\faGithub~\footnotesize\url{https://github.com/Pleias/toxic-commons} \normalsize

Model: \raisebox{-2.2pt}{\includegraphics[scale=0.09]{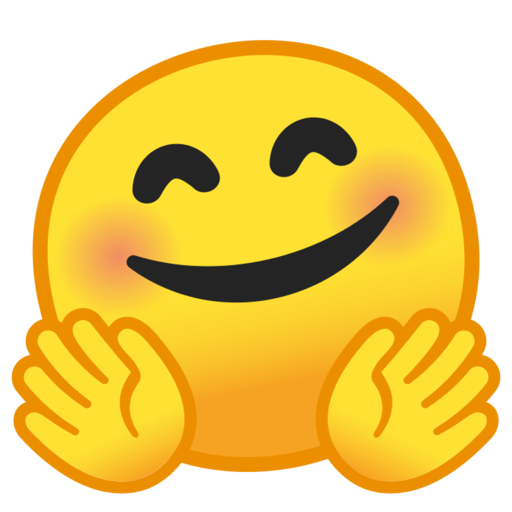}}
\href{https://huggingface.co/PleIAs/celadon}{\texttt{Celadon}} ~~~ Dataset: \raisebox{-2.2pt}{\includegraphics[scale=0.09]{figures/huggingface.png}} \href{https://huggingface.co/datasets/PleIAs/ToxicCommons}{\texttt{ToxicCommons}} 
\end{center}

\section{Introduction}
There are growing calls for transparency surrounding Large Language Models (LLM) \citep{bommasani2024foundation}. Legislation about safety and openness requirements such as EU AI Act \citep{eu_ai_act} and Executive Order on the Safe, Secure, and Trustworthy Development and Use of Artificial Intelligence \citep{executive_order_ai} are the first steps made by governments to ensure some form of LLM transparency. As emphasis on open models increases, developers open more parts of their training pipelines \citep{liu2023llm360,mehta2024openelm,faysse2024croissantllm}. The performance gap between open and closed-weights models is closing fast: Meta's Llama 3.1 405B Instruct \citep{llama3} and Mistral's Mistral Large 2 \citep{mistral2} are both competitive when compared with frontier closed-source models \citep{chiang2024chatbot}. Despite the increasing popularity and power of open-weight models, many modern ``open" models do not meet the requirements of the openness matrix \citep{liesenfeld2024rethinking}, where the openness of a model takes into consideration not just model weights, but model-related artifacts such as training code and training data, documentation, and access to the model.
Open training data is the dimension of openness that lags behind the most \citep{bommasani2024foundation}. Since some of the closed models are clearly trained on proprietary data that is scraped without the consent of the data creator, the lack of openness surrounding model training data poses significant ethical challenges \citep{longpre2024consent}. Open data presents unique challenges that this work tries to address. We hope that the provided data, pipeline and filtration principles will help to intensify work on open-data LLMs. 

\subsection{Curating Public Domain Data}
Closed-source model developers consistently utilize copyrighted material in their training sets. OpenAI has been sued for using copyrighted newspaper data \citep{RobertsonNYTimes}, while Apple, Nvidia, and Anthropic have all been criticized by the YouTube community after being caught using copyrighted videos in their models \citep{GilbertsonReisner}. In contrast, the open-source community is pushing for greater transparency by adopting open-data practices. Key leaders have released not only open-weights models but training data as well, such as Dolma \citep{soldaini2024dolma} used to train OLMo \citep{groeneveld2024olmo} and the Pile \citep{gao2020pile} used to train the Pythia models \citep{biderman2023pythia}, \textit{inter alia}. 

Upon inspection, these open corpora contain a lot of implicit bias and stereotypes against certain gender, racial, ethnic, or disability groups \citep{bender2021dangers, dodge-etal-2021-documenting, luccioni-viviano-2021-whats}. Unfortunately, the documentation of the open data curation pipelines is often lacking, and methods can be \textit{ad hoc} or have little regard for the context of the data \citep{henderson_2022_pile_law}. More resources need to be dedicated to curating and documenting the construction of these datasets in order to reduce disproportionate impacts on the most vulnerable groups at the margins of society \citep{birhane2019algorithmic, birhane2021large, bender2021dangers}. Data curation, as well as other methods designed to address safety of open LLMs are to be understood in Bayesian sense: though it is highly unlikely that one could ever design a safety tool that could enable non-toxic generation with probability one, every effort reduces the prior of the model for toxic behavior.

Common Corpus\footnote{\url{https://huggingface.co/collections/PleIAs/common-corpus-65d46e3ea3980fdcd66a5613}} was recently released. It is the largest public domain dataset for training language models. It comprises multilingual datasets, mostly coming from books and periodicals, and as such has a great deal of cultural heritage content. In order to be in the public domain, many of these texts are historical. Most of the texts come from the 18th and 19th centuries, and some are much older.

This kind of public domain data poses a different set of challenges, relative to existing open datasets, which are often largely comprised of web text. While the toxicity of web text is well documented, \citep{bender2021dangers, groeneveld2024olmo}, we know much less about the risks posed by historical data. As societal norms, especially in the language surrounding minoritized and racialized groups, have changed substantially even over the last few decades. In order to use historical text data safely, we propose that it be classified and filtered. 

Existing toxicity classifiers are not well-suited to this task. Toxicity classifiers perform poorly on out-of-domain data \citep{adragna2020fairness}. Existing toxicity classifiers are trained to perform well on web text, which is unlike Common Corpus in style and content, due to text domain and genre, language change, and other factors. Toxicity classifiers have been shown to perform poorly on historical data \citep{borenstein-etal-2023-measuring}. Additionally, much of the data in Common Corpus is collected via optical character recognition (OCR), which introduces noise into the data. OCR affects many NLP tasks \citep{van2020assessing}. Specifically, \citet{borenstein-etal-2023-measuring} show that OCR errors lead to less accurate toxicity classification. We test an existing toxicity classifier on a sample from Common Corpus. As it was trained to detect toxicity in contemporary web text, it generalizes very poorly to Common Corpus data. For example, we find that the toxicity classifier tends to assign higher toxicity when texts contain large amounts of capital letters. The classifier we tested also seems overly sensitive to words that may often appear in toxic or violent content, but are not necessarily toxic or harmful in many contexts, e.g. `fear', `army'. We provide details and examples in Appendix A

Finally, there is less data available to developers in the public domain. While web text datasets are comprised of hundreds of gigabytes of data, historical data which has been digitized is much more sparse. Using ill-suited filtering methods may lead to removing large amounts of already limited data. We must not only identify problematic content with high accuracy, but we must also find a way to transform data with low-levels of toxicity into useful data that contributes to a safer open model.

In this paper, we create a classifier specific to Common Corpus in order to filter the data and prepare it for synthetic treatment. We also propose this as a framework, which can be used in combination with our open and replicable methods to be adapted for any dataset. We show that it is possible to efficiently and effectively curate datasets that can help reduce unwanted model behaviors. 

\section{Related Work}

\subsection{Safety Guardrails}

Much of the work on creating safer language models focuses on post-training procedures. For models deployed through APIs and web interfaces, continual refinement and tuning – both through parameter updates and also through system prompting and output moderation – are common approaches. These methods have clear advantages over open-weights models when it comes to trust and safety; however, this does not always translate to explicitly better performance in this respect, as even frontier versions of these models can be jail-broken with ease \citep{wei2023jailbrokendoesllmsafety, wu2024newerallmsecurity}. 

There are also existing tools which can be deployed by developers of generative AI software to prevent AI-based software from violating safety policies. For example, Meta released Llama Guard 3, Prompt Guard, and Code Shield \citep{llama3} in order to help developers utilizing generative AI tools ensure that model outputs adhere to safety policies. Code Shield as a tool can be deployed at inference time, providing obvious benefits to developers of closed-source software. Google's recent Shield Gemma \citep{zeng2024shieldgemmagenerativeaicontent} builds upon previous success of Llama Guard 3, and helps the open-source community by advancing content moderation tools. However, when it comes to open-weights models, safety measures can only be implemented during pre-training and fine-tuning, and cannot be deployed in between the model and the user. Serving as an example of the difficulties open-weights model developers face, Llama Guard 3 was integrated into the Llama 3.1 release, and yet the results of this effort were quickly undone \citep{gade2023badllama}. As a result, we argue that safety pipelines for open-source generative AI must start with thorough data filtering.

\subsection{Filtering Model Training Data} 
Apart from post-training procedures, a strategy for mitigating harmful model behaviors is dataset curation, often through filtering out harmful content in the training data. Most toxicity filtering on pre-training datasets falls into one of three categories: URL-based filtering, lexicon-based filtering, and classifier-based methods \citep{albalak2024survey}. Datasets like FineWeb \citep{penedo2024fineweb} and the Llama 3 pre-training dataset \citep{llama3} filter out content that comes from URLs that are on blacklists, such as the UT1 URL Blacklist \citep{url_blacklist}. While this method may effectively reduce the amount of pornographic content, it does not directly address other kinds of toxic or offensive content that frequently appears in web text. It has also been shown that blocking pornographic content also reduce the amount of LGBTQ+ and African American English (AAE) text \citep{dodge-etal-2021-documenting, zhou-etal-2021-challenges}. This in turn may lead to a lack of diversity or representation. 

Lexical filtering refers to removing documents that contain blacklisted words. This method was used to create the Colossal Clean Crawled Corpus (C4; \citet{raffell2020c4}), which is used to train Gopher \citep{rae2021scaling}, Chinchilla \citep{chinchilla}, T5 \citep{raffell2020c4}, and more. In this case, any document that contained any word from the ``List of Dirty, Naughty, Obscene, and Otherwise Bad Words"\footnote{\url{https://github.com/LDNOOBW/List-of-Dirty-Naughty-Obscene-and-Otherwise-Bad-Words}} was removed. 
The ROOTS corpus \citep{roots_corpus}, which was used to train BLOOM \citep{bloom} also filtered out texts that contained high ratios of words associated with pornographic spam. 
Aggressive filtering may lead to performance degradation \citep{henderson_2022_pile_law}, because filters remove high quality data \citep{longpre-etal-2024-pretrainers}. This can occur especially when blacklisted words are filtered without considering their context, as some slurs or offensive terms may be acceptable in some specific contexts \citep{warner-hirschberg-2012-detecting}. For example, a slur or offensive term may be used in the discussion of that term without being toxic or offensive \citep{henderson_2022_pile_law}. Additionally, some words in the commonly used ``bad words" lists have other, non-offensive uses. 
For example, we found high instances of `bitch' used in one dataset of historical newspapers, which always referred to female dogs or other animals. 

Finally, there are classifier-based methods for dataset filtering. When toxicity classifiers have been used to filter pre-training data, the classifiers have often been black-box, proprietary products. Gopher data was filtered using Google SafeSearch \citep{rae2021scaling}, which primarily filters out pornographic content \citep{monea2023know}. Palm 2 used a fixed version of Perspective API\footnote{\url{https://perspectiveapi.com/}} to generate control tokens for a subset of the data \citep{anil2023palm}. The Perspective API specifically has been shown to be more accurate than heuristics or rule-based classifiers \citep{longpre-etal-2024-pretrainers}, but black-box APIs are not reproducible \citep{pozzobon-etal-2023-challenges}. The Dolma dataset was filtered using FastText classifiers trained to identify text using the labels ``hate" and ``NSFW" \citep{soldaini2024dolma}. 
To the best of our knowledge, this work presents one of the only large-scale filtering procedures using open or reproducible methods for open pre-training data. There have been limited studies on the effects of these filters, however, more research is needed to understand how they affect model behavior \citep{albalak2024survey}. 

\section{Data Filtering Pipeline}
In this paper, we propose a strategy of safety guardrails for open-data language modeling, especially well suited for issues specific to public domain data. We release a classifier and dataset specifically created for toxicity filtration in open-source datasets. In this section, we outline our strategy for the curation of the dataset and creation of the classifier:
\begin{enumerate}
    \item \textbf{Annotation Process}: We use human annotations to guide the LLM annotation process, in order to allow for scalability of annotation while remaining true to the ground truth of the samples.
    \item \textbf{Toxicity Classifier}: Using the LLM-generated annotations, we train a classifier capable of identifying toxic content across four levels of severity (0, 1, 2, 3) in five different categories (race and origin-based, gender-based, ability-based, religion-based discrimination, as well as violence). 
    \item \textbf{Synthetic Re-writing For Content Moderation}: After classifying texts based on the above categories, we implement different content moderation policies depending on the severity of the problem. Texts with only low levels of toxicity are included with a content warning, while the most egregious samples get re-written and reintroduced into the corpus in a ``detoxified" form. 
\end{enumerate}

We recognize that the dataset, composed of historical and post-OCR data, poses unique challenges to general open-source data, let alone closed-source data. It is our hope that not only can our dataset and classifier be used on similar types of data, but our process can be generalized to other use cases by the community at large. 

\section{Dataset}
We chose 13 cultural heritage datasets from Common Corpus, which includes datasets from nine languages. 
All datasets are comprised of copyright-free, public domain text data. Several datasets were created from OCR data, so we applied OCR correction using OCRonos\footnote{\url{https://huggingface.co/PleIAs/OCRonos}} to reduce noise in the data. OCR correction does not remove all OCR errors, so there is still some noise from the digitization process in the data. 

\begin{table}[h!]
\centering
\begin{tabular}{l|c}
    \textbf{Language} & \textbf{\# Samples}  \\ \hline
    \href{https://huggingface.co/datasets/PleIAs/US-PD-Newspapers}{US PD Newspapers} &  250k \\
    \href{https://huggingface.co/datasets/storytracer/US-PD-Books}{US PD Books} &  250k \\
    \href{https://huggingface.co/datasets/PleIAs/French-PD-Newspapers}{French PD Newspapers} & 250k \\
    \href{https://huggingface.co/datasets/PleIAs/French-PD-Books}{French PD Books} & 250k \\
    \href{https://huggingface.co/datasets/PleIAs/German-PD}{German PD} &  100k \\
    \href{https://huggingface.co/datasets/storytracer/German-PD-Newspapers}{German PD Newspapers} & 100k \\
    \href{https://huggingface.co/datasets/PleIAs/Italian-PD}{Italian PD} & 200k \\
    \href{https://huggingface.co/datasets/PleIAs/Spanish-PD-Books}{Spanish PD Books} & 100k \\
    \href{https://huggingface.co/datasets/PleIAs/Spanish-PD-Newspapers}{Spanish PD Newspapers} & 100k \\
    \href{https://huggingface.co/datasets/PleIAs/Polish-PD}{Polish PD} & 100k \\
    \href{https://huggingface.co/datasets/PleIAs/Portuguese-PD}{Portuguese PD} & 100k \\
    \href{https://huggingface.co/datasets/PleIAs/Dutch-PD}{Dutch PD} & 100k \\
    \href{https://huggingface.co/datasets/PleIAs/Latin-PD}{Latin PD} & 100k \\
\end{tabular}
    \caption{Number of LLM-annotated samples from each dataset.}
    \label{tab:offensive_word_count}
\end{table}

\section{Annotation Process}
The annotation process can be further subdivided into two distinct steps: human annotations and LLM annotations.  

\subsection{Human Annotations}

In order to create a baseline to compare the LLM annotations against, human-annotation of a small subset of 15 chunks was necessary\footnote{We provide these annotations in full in Appendix B.}. These annotations were on a mix of low- and high-toxicity datasets to monitor over- and under-sensitivity. The annotators independently annotated and scored ten different texts, before creating a unified ground-truth dataset from these annotated examples to evaluate LLM performance. 

We identify five types of bias to train the classifier to detect, which are frequently identified in the literature: 

\begin{itemize}
    \item[] \textbf{Race and origin-based bias}: includes racism as well as bias against someone's country or region of origin or immigration status, especially immigrant or refugee status \citep{zhang2023mitigating}.
    \item[] \textbf{Gender and sexuality-based bias}: includes sexism and misogyny, homophobia, transphobia, and sexual harassment \citep{barronqueerbench}.
    \item[] \textbf{Religious bias}: any bias or stereotype based on someone's religion.
    \item[] \textbf{Ability bias}: bias according to someone's physical, mental, or intellectual ability or disability \citep{farrelly2023current}.
    \item[] \textbf{Violence and abuse}: overly graphic descriptions of violence, threats of violence, or calls or incitement of violence \citep{saha2023blp}.
    
\end{itemize}

The annotators rated each text on a dimension of bias on a scale from 0 to 3. This scale was used for two reasons:
it is easily explainable, such that we can make clear distinctions between categories, and it should not introduce any further complexities into the classification process. Any scale with more than four classes would prove to be difficult to properly explain to a language model. LLM annotators are most effective when decisions are left up to as little interpretation as possible, and we felt that by clearly defining these four categories, we would attain the most success. Many classifiers are either binary, toxic or not toxic, or on a continuous scale between 0 and 1. Neither of these methods suit our task; a binary scale cannot properly capture the nuances of our dataset, and a continuous scale is impossible to curate using LLM annotators.

The four categories for classifying each dimension of bias are provided below:
\begin{enumerate}
    \item[0:] Text mentions stereotypes about a particular group of people or acts of violence, but does so in an objective way; text does not promote biases against any group or acts of violence.
    \item[1:] Text implies biased beliefs against a group of people or implicitly makes suggestions of violent acts, but does not do so explicitly.
    \item[2:] Text clearly expresses biased beliefs against a group of people, but does not do so in an overly aggressive manner, nor does it actively promote the spread of such beliefs. 
    \item[3:] Text overtly promotes hatred towards or disenfranchisement of a group of people or promotes the superiority of one group of people over another; text calls for violence against an individual or group of people; text persuasively describes stereotypes or biases against a group of people or characteristics associated with a particular group (especially groups according to race, sex, gender, origin, ability, or religion). 
\end{enumerate}

As texts were rated, the annotators did not consider any slurs or offensive terms in their annotations to reflect the process of the LLM annotations. The intention was to develop an annotation process that would detect implicit bias but would not be overly sensitive to certain lexical items.

\subsection{LLM Annotations}

We tested three models and compared zero-shot and few-shot prompting strategies for the LLM annotations. We tested GPT-3.5, Claude 3 Haiku, and Llama 3.1. 
Claude was most likely to refuse to provide an annotation. Claude also tended to give higher ratings -- often giving a score of 3 on a given dimension, which often did not align with the human annotations. 
GPT-3.5 was more likely than the other models to underestimate the human toxicity scores. 
We found that Llama 3.1 was well-aligned with human annotations. Additionally Llama 3.1 is an open-weight model, which makes this procedure more easily replicable.
We did not find that few-shot prompting improved the alignment of the model annotations with the gold human annotations. Few-shot prompting tended to make the models give higher toxicity ratings, even when not appropriate to do so. We report comparisons and examples in Appendix E.

Using the human annotations as a baseline, we refined a prompt to generate scores and annotations on each of the five dimensions of bias. The prompt uses the same scale from 0 to 3 as for the human annotations. The prompt was designed to have a relatively high bar for toxicity, so that only the most offensive content had a score of 3 for a given dimension. We report the prompt we developed in Appendix C. 
We provide examples of annotations for toxic text in Appendix F. Note that the content of the texts and annotations may be offensive or upsetting. 
We release the LLM-annotated dataset, ToxicCommons\footnote{\url{https://huggingface.co/datasets/PleIAs/ToxicCommons}}, with 2M samples.

\begin{figure}[h!]
    \centering
    \includegraphics[width=0.95\linewidth]{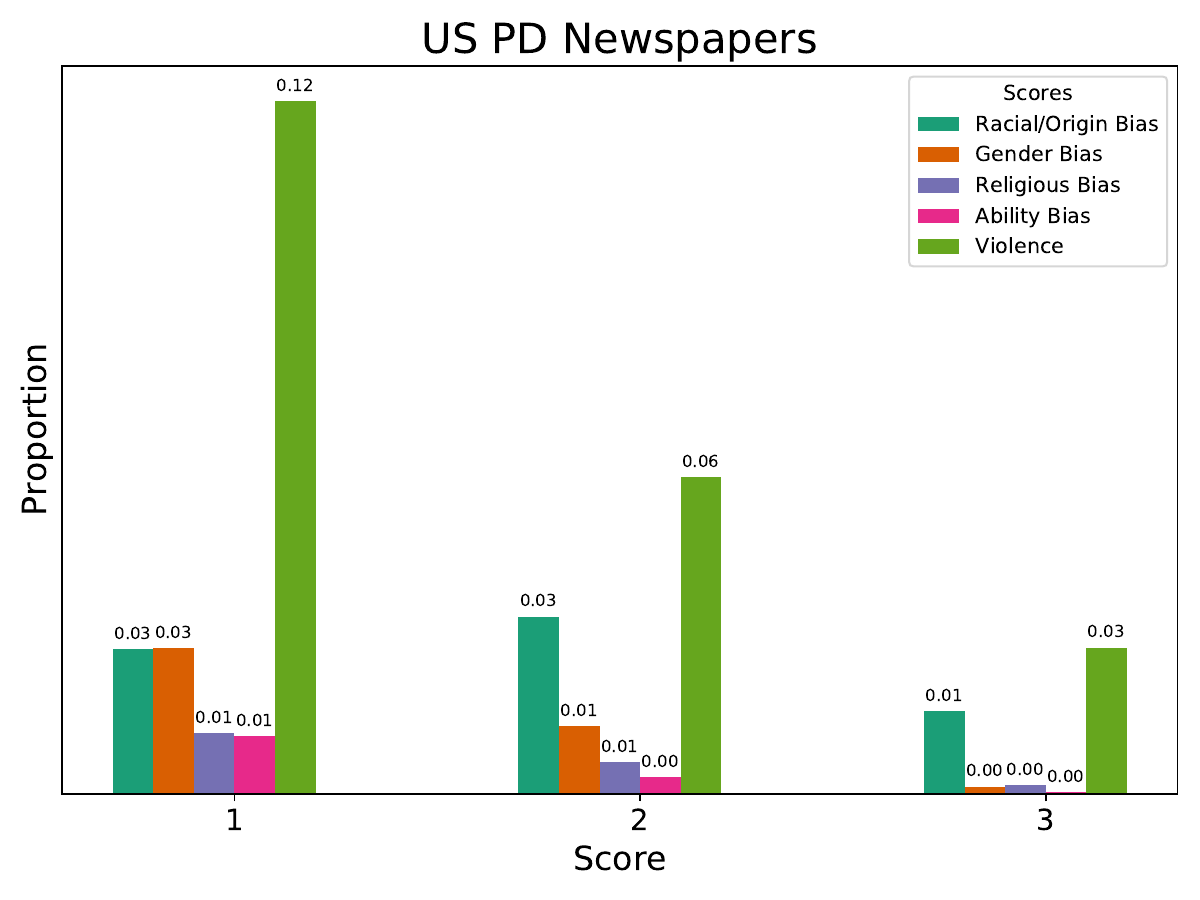}
    \caption{Distribution of toxicity scores 1-3 for each of the five dimensions for US PD Newspapers. Not shown are the counts of texts with scores of 0, which represented the majority of texts in the corpus.}
    \label{fig:us_newspaper_scores1}
\end{figure}

We report the rates of each score above 0 for each dataset. Figure \ref{fig:us_newspaper_scores1} shows the scores for the US Newspapers dataset. The scores for the other datasets are reported in Appendix D. We find that the dimension with the highest rates of scores above 0 is Violence. Llama 3.1, as well as other models we tested, were generally very sensitive to any description of violent events. For the other dimensions, we find that generally less than 5\% of texts are assigned a score above 0 for each of the other dimensions.

Across the different datasets, US Newspapers, US Books, French Books, and Italian datasets have the highest toxicity scores, especially for violence and especially for the lowest toxicity level (1). The books datasets have the highest religious bias scores, which may be because books were more likely to have any religious content. High toxicity (3) scores are relatively consistent across datasets, always between 0-3\% of texts. Across all datasets, we found highest toxicity rates for the violence and abuse dimension, relative to other dimensions. When tested different models, we found that all models were the most sensitive to the violence dimension.

\section{Toxicity Classifier}
We use the LLM annotations to train Celadon\footnote{\url{https://huggingface.co/PleIAs/celadon}}, our toxicity classifier. The model was trained on just over 640K samples of data taken from the LLM annotation dataset, and validated on around 133K samples. The entire LLM annotation dataset comprises 2 million samples across the different datasets; however, due to the overwhelming majority of non-toxic data in the data, we did not use the whole dataset. Table \ref{tab:training_dist} shows the number of samples, per category, per class, for the training split. An additional challenge was the much smaller quantity of non-zero-score samples for the ability-based discrimination dimension. Had we downsampled the majority classes completely to match the minority classes, the classifier would not have been trained on data that was truly representative of the distribution of samples in Common Corpus. Instead, we chose to make the maximum number of not toxic samples (samples where the sum of toxicity scores was 0) equal to the sum of all of the samples with at least some toxicity (samples where the sum of toxicity scores was 1 or more). Additional training details are provided in Appendix G, including the score distribution for the validation and training sets. Training code is available on GitHub\footnote{\url{https://github.com/Pleias/toxic-commons}}.

\begin{table}[h!]
\centering
\begin{tabular}{lcccc}
\hline
\multicolumn{5}{c}{Training Data Distribution} \\
\hline
            Score     & 0     & 1    & 2    & 3    \\
\hline
Racial/Origin & 0.934  & 0.021  & 0.028 & 0.017  \\
Gender/Sex    & 0.927  & 0.046  & 0.025  & 0.002   \\
Religion      & 0.878  & 0.054  & 0.054  & 0.014  \\
Ability       & 0.979  & 0.016  & 0.004   & 0.001    \\
Violence            & 0.638  & 0.210 & 0.107 & 0.046 \\
\hline
\end{tabular}
\caption{Proportion of scores for each dimension out of 640,808 total samples.}
\label{tab:training_dist}
\end{table}

\subsection{Model Architecture}
Celadon is a 5-headed classifier with a DeBERTa-v3-small \citep{he2021debertav3} base. Each head is a separate linear layer, which utilizes weights from a custom weighted loss function in order to produce the final logits. Below is the weighted cross entropy loss function:

\[
\mathcal{L}_{\text{WCE}}(x, y) = - \sum_{i=1}^{C} w_i \cdot y_i \cdot \log(\hat{y}_i)
\]

Where:
\begin{itemize}
    \item \( x \) represents the input logits, and \( y \) represents the true class labels.
    \item \( \hat{y}_i \) is the predicted probability for class \( i \), computed using the softmax function.
    \item \( C \) is the total number of classes.
    \item \( w_i \) is the weight assigned to class \( i \).
\end{itemize}

The weights \( w_i \) are updated dynamically based on the confusion matrix. 

\begin{itemize}
    \item \( \mathbf{CM}_{ij} \) is the number of instances of class \( i \) predicted as class \( j \),
    \item The row sums \( \text{RowSum}_i = \sum_{j=1}^{C} \mathbf{CM}_{ij} \) represent the total number of samples in the true class \( i \),
    \item The column sums \( \text{ColSum}_j = \sum_{i=1}^{C} \mathbf{CM}_{ij} \) represent the total number of samples predicted as class \( j \),
    \item The diagonal \( \mathbf{CM}_{ii} \) represents the correctly classified samples for class \( i \),
\end{itemize}

We compute the false negatives (FN) and false positives (FP) for each class:

\[
\text{FN}_i = \text{RowSum}_i - \mathbf{CM}_{ii}
\]
\[
\text{FP}_i = \text{ColSum}_i - \mathbf{CM}_{ii}
\]

The total error for class \( i \) is the sum of false negatives and false positives:

\[
\text{TotalError}_i = \text{FN}_i + \text{FP}_i
\]

To update the weights for each class, we apply a logarithmic scaling:

\[
w_i = \frac{\log(1 + \text{TotalError}_i)}{\sum_{k=1}^{C} \log(1 + \text{TotalError}_k)} \times C
\]

\subsection{Evaluation Results}
Classifier performance was further tested on a held-out sample of 133K LLM annotations. Tables \ref{tab:race_results}, \ref{tab:gender_results}, \ref{tab:religion_results}, \ref{tab:ability_results}, and \ref{tab:violence_results} outline the test-set performance of the classifier on each of the various categories. 
We found very high performance according to precision, recall, F1, and accuracy metrics; however, we argue that these metrics are misleading for this task. Due to the low prevalence of ability-based discrimination, the classifier performs extremely well on all metrics by predicting the majority class (0) the most often. As a result, we included weighted accuracy as an important metric for model evaluation, given the distribution of the scores across all four classes for each category. 

The classifier performs the best in terms of weighted accuracy for the violence category, yielding a weighted accuracy of 74\%. The model has lower values of precision, recall, F1, and accuracy for this category, relative tother. We argue this is an indicator of good model performance, as it suggests that the classifier is correctly predicting at or around the correct class most of the time. 
Specifically, we note the distribution of classifications per category. For religion-based toxicity, and violence, the distribution of guesses centers about the correct class. Critically, samples that are true 0's are likeliest to be predicted as a 0, with a 1 being next likeliest at a much lower rate. True 1's are likeliest to be predicted 1's, then 0's, then 2's, 2's are likeliest to be predicted as 2's, 1's, then 3's, and 3's are likeliest to be predicted as 3's or 2's. 
This is an important feature of the classifier, as we aim not to be over-sentitive in the filtering of this dataset.
Furthermore, the classifier consistently predicts one class lower than the true class as the second majority class, which signals that the classifier does not often to predict higher scores than is true. The purpose of the classifier is to quickly and efficiently identify samples of text which should be annotated by an expert LLM annotator, and for this task, it excels in its duties.  

\begin{table}[h!]
\centering
\begin{tabular}{l|p{0.8cm}|p{0.8cm}|p{0.8cm}|c|p{0.8cm}}
\multicolumn{6}{c}{\textbf{Race/Origin-Based Discrimination Results}} \\
\hline
           & Prec. & Recall & F1 & Acc. & WA \\
\hline
Value      & 0.953    & 0.951 & 0.952   & 0.951   & 0.734            \\
\end{tabular}
\caption{Test Set Accuracy Measures: Precision, Recall, F1, Accuracy, and Weighted Accuracy (WA)}
\label{tab:race_results}
\end{table}

\begin{table}[h!]
\centering
\begin{tabular}{l|p{0.8cm}|p{0.8cm}|p{0.8cm}|c|p{0.8cm}}
\multicolumn{6}{c}{\textbf{Race/Origin-Based Discrimination Results}} \\
\hline
           & Prec. & Recall & F1 & Acc. & WA \\
\hline
Value      & 0.957    & 0.955 & 0.956   & 0.955   & 0.714            \\
\end{tabular}
\caption{Test Set Accuracy Measures: Precision, Recall, F1, Accuracy, and Weighted Accuracy (WA)}
\label{tab:gender_results}
\end{table}

\begin{table}[h!]
\centering
\begin{tabular}{l|p{0.8cm}|p{0.8cm}|p{0.8cm}|c|p{0.8cm}}
\multicolumn{6}{c}{\textbf{Race/Origin-Based Discrimination Results}} \\
\hline
           & Prec. & Recall & F1 & Acc. & WA \\
\hline
Value      & 0.934    & 0.931 & 0.935   & 0.931   & 0.729            \\
\end{tabular}
\caption{Test Set Accuracy Measures: Precision, Recall, F1, Accuracy, and Weighted Accuracy (WA)}
\label{tab:religion_results}
\end{table}

\begin{table}[h!]
\centering
\begin{tabular}{l|p{0.8cm}|p{0.8cm}|p{0.8cm}|c|p{0.8cm}}
\multicolumn{6}{c}{\textbf{Race/Origin-Based Discrimination Results}} \\
\hline
           & Prec. & Recall & F1 & Acc. & WA \\
\hline
Value      & 0.985    & 0.985 & 0.985   & 0.985   & 0.697            \\
\end{tabular}
\caption{Test Set Accuracy Measures: Precision, Recall, F1, Accuracy, and Weighted Accuracy (WA)}
\label{tab:ability_results}
\end{table}

\begin{table}[h!]
\centering
\begin{tabular}{l|p{0.8cm}|p{0.8cm}|p{0.8cm}|c|p{0.8cm}}
\multicolumn{6}{c}{\textbf{Race/Origin-Based Discrimination Results}} \\
\hline
           & Prec. & Recall & F1 & Acc. & WA \\
\hline
Value      & 0.819    & 0.799 & 0.806   & 0.799   & 0.745            \\
\end{tabular}
\caption{Test Set Accuracy Measures: Precision, Recall, F1, Accuracy, and Weighted Accuracy (WA)}
\label{tab:violence_results}
\end{table}

\section{Pre-Training Data Curation}
While the classifier allows us to perform actions based on the toxicity levels of a piece of text, we believe that the data processing decisions are an important contribution of this paper. Specifically, given the classes, we make three decisions based on the classified levels of toxicity:

\begin{enumerate}
  \item \textbf{No Toxicity:} A total score of 0-3 with no individual category score exceeding a 2. These texts either have no toxic content, or very low levels not warranting any further treatment. We allow some breathing room with scores of 2 for any overly-sensitive cases. 
  \item \textbf{Mild Toxicity:} A total score of between 4 and 6, or a total score of 3 where that score comes from a 3 in a single category. For a lower score of 3, we require the sole contribution to come from a single category, while higher scores indicate toxicity across multiple categories in a significant manner. 
  \item \textbf{Toxic Content:} The sum of scores is 7 or greater. This text likely contains toxic content that requires scrutiny from an LLM annotator. 
\end{enumerate}

The thresholds used to divide the data into these three categories reflect our desire to not over-correct texts in our dataset. As mentioned previously, over-correction is almost as detrimental to our work as under-correction. It is critical that the historical nature and context of the data be preserved as much as possible, and removing too much content would reduce many of the benefits of working with public domain -- and especially cultural heritage -- data. As a result, we aimed to be more lenient with our thresholds, accounting for the fact that there are two potential vectors for over-sensitivity: the LLM annotations, and then the distilled classifier. These thresholds reflect deliberate decisions that we made as a result of our specific task, but it is important to note that as we release this pipeline to the broader community, these thresholds will adjust depending on individual needs. The classifier acts as a useful intermediary: the smaller architecture of the DeBERTa-v3-small model means that it is deployable at scale much more efficiently than a LLM\footnote{We found that Celadon is over 40 times faster than Llama for annotations. On an NVIDIA H100 GPU, it took 3.43 hours (205 minutes) to annotate 100k samples with Llama, and only 5 minutes with Celadon. This is a significant speed-up, when considering annotating trillions of tokens of text.}. 
However, we reintroduce the LLM back into the pipeline for the final step of the process, which annotating or re-writing texts which have been labeled as toxic. The classifier proves to be an incredible useful heuristic when it comes to identifying toxic text, however, we recognize due to the accuracy-efficiency trade-off, it is not always accurate. As a result, the re-introduction of the LLM annotator is a critical part of the pipeline, both for annotating or re-writing text that is actually problematic, or to better contextualize text that isn't. 

\textbf{Content Warning.} The first LLM treatment for toxic data is to synthetically generate a content warning for the text. The intention is to develop datasets that can help models to reason about harmful content and ideas. To do this, we use Llama 3.1. We wrote the initial prompt in English, and then translated it to the remaining languages, in order to ensure that the LLM annotator would not default to English for its annotations, and instead match the parent language of the original text. Were we to obtain results where the entirety of the content warnings were in English, we would not only introduce a greater imbalance of English tokens into the dataset, but we would also artificially align content moderation outputs with the English language. Further, by utilizing Llama 3.1, we rely on its multilingual capabilities in order to better contextualize the texts in different languages. We provide the LLM annotator with the previously defined bias categories as well as their definitions. We then prompt it to create a content warning that includes justification for why the content may be toxic, utilizing cultural, ethical, legal, or moral arguments. We also encourage it to state if the provided text is not toxic in any manner. In this vein, it is crucial that we do not provide the axes upon which the classifier has deemed the text toxic, so as to not skew the thoughts of the LLM annotator. Should it deem a provided piece of text not toxic across any axis, this additional content warning will still be kept, as it provides additional alignment benefits for the final model.  We report the prompts used for the content warning in Appendix H. 

\textbf{Synthetic Re-writing.} We follow a similar process as with generating the content warnings to synthetically re-write content that is extremely toxic.  We writing the prompt in English and then translate it to the other languages. We also provide the LLM with the categories of bias and their definitions, and we prompt it to provide the re-written text, reworded in order to remove any traces of toxicity, while staying as authentic as possible to the original style, tone, context, and word choice of the provided text. We also ask it for a list of changes made to the original text in order to better engineer annotations that adhere to the guidelines provided. As with the content warning, should the LLM decide that the provided text is not toxic across any of the categories, it will simply return the provided text, along with the reasoning that it is not toxic. The most interesting results from this process occur when the LLM annotator is fed truly toxic content, that contains no meaningful language that isn't a part of representing a toxic belief. In this case, the annotator will re-write the entire text in an anti-toxic manner. We report the prompts used for the synthetic re-writing in Appendix I.

\section{Discussion and Conclusion}
Due to the nature of public domain data for LLM training that contrasts with closed data, we demonstrate a need to create a new approach that properly balances the desire to better align language models trained with the dataset with the importance to preserve the inherent properties of the text data. 
One approach is to filter out everything that is at all toxic. This is possible to do with larger web-based or otherwise open but not public domain data, and still remain with enough tokens to train an LLM. However, we are more limited in the curation of open and public domain data that is also relatively free of harmful content. We propose a compute- and data-efficient approach that preserves the vast majority of the data, while still mitigating harms.

We release this pipeline, our annotated dataset, the classifier, as well as the prompts used for synthetic data generation from toxic text sources. To our knowledge, this pipeline represents the an effective method of handling toxic content in the pre-training phase on public domain data, and has the potential for broader applications across the open source AI community.

\section{Limitations}
We recognize that our methods come with their limitations. First, our corpus is not that linguistically diverse, given that we use mostly Western European languages. Even further, there are major differences in terms of the amount of resources available for the different languages – offensive word lists, data available, and benchmarks are all lacking even for our relatively popular languages.
Second, past our initial human-annotated data, we were financially incapable of paying human annotators to verify all of the annotations. Current state-of-the-art for this type of data curation often uses human annotators or verifiers in order to ensure the soundness of the training labels. While this potentially could improve accuracy, it also exposes humans to toxic text that is taxing to work with and difficult to read. By utilizing LLMs for the majority of the annotation process, we remove any human exposure to this content, with the risk of slightly less accurate annotations. Finally, in releasing our dataset to the public, we are knowingly publishing some content which is toxic and harmful in nature. We acknowledge that there are certain individuals who may use the data in our dataset to perpetuate bias across any of the five toxicity categories outlined in our paper, but as staunch proponents of full transparency and openness, we firmly believe that the greater benefits to the community that come from applied use cases will outweigh the costs of this decision.

\section{Acknowledgments}

The authors acknowledge the resource by Jean Zay (compute grant \#GC011015451).
The authors would also like to thank the other members of the PleIAs Research team for helpful discussion. 

\bibliographystyle{apalike}
\bibliography{toxicity}

\section{Reproucibility Checklist}

This paper:

\begin{enumerate}
    \item Includes a conceptual outline and/or pseudocode description of AI methods introduced: 
    \item Clearly delineates statements that are opinions, hypothesis, and speculation from objective facts and results: yes 
    \item Provides well marked pedagogical references for less-familiar readers to gain background necessary to replicate the paper: yes 
\end{enumerate}

Does this paper make theoretical contributions?: no.

Does this paper rely on one or more datasets?: yes. 
\begin{enumerate}
    \item A motivation is given for why the experiments are conducted on the selected datasets: yes.
    \item All novel datasets introduced in this paper are included in a data appendix: yes.
    \item All novel datasets introduced in this paper will be made publicly available upon publication of the paper with a license that allows free usage for research purposes: yes
    \item All datasets drawn from the existing literature (potentially including authors’ own previously published work) are accompanied by appropriate citations: yes.
    \item All datasets that are not publicly available are described in detail, with explanation why publicly available alternatives are not scientifically satisfying: N/A
\end{enumerate}

Does this paper include computational experiments?: yes

\begin{enumerate}
    \item Any code required for pre-processing data is included in the appendix: yes
    \item All source code required for conducting and analyzing the experiments is included in a code appendix: yes
    \item All source code required for conducting and analyzing the experiments will be made publicly available upon publication of the paper with a license that allows free usage for research purposes: yes
    \item  All source code implementing new methods have comments detailing the implementation, with references to the paper where each step comes from: yes
    \item If an algorithm depends on randomness, then the method used for setting seeds is described in a way sufficient to allow replication of results: N/A
    \item This paper specifies the computing infrastructure used for running experiments (hardware and software), including GPU/CPU models; amount of memory; operating system; names and versions of relevant software libraries and frameworks: yes
    \item This paper formally describes evaluation metrics used and explains the motivation for choosing these metrics: yes
    \item This paper states the number of algorithm runs used to compute each reported result: yes
    \item: Analysis of experiments goes beyond single-dimensional summaries of performance (e.g., average; median) to include measures of variation, confidence, or other distributional information: yes
    \item The significance of any improvement or decrease in performance is judged using appropriate statistical tests (e.g., Wilcoxon signed-rank): yes
    \item This paper lists all final (hyper-)parameters used for each model/algorithm in the paper’s experiments: yes
    \item This paper states the number and range of values tried per (hyper-) parameter during development of the paper, along with the criterion used for selecting the final parameter setting: yes
\end{enumerate}

\clearpage

\appendix

\bigskip
\newpage
\onecolumn

\renewcommand\thesection{\Alph{section}}
\renewcommand\thefigure{\Alph{section}.\arabic{figure}}
\renewcommand\thetable{\Alph{section}.\arabic{table}}
\counterwithin{figure}{section}
\counterwithin{table}{section}

\begin{center}
    \large{\textcolor{red}{\textbf{CONTENT WARNING: Some of the text below may offensive or graphic in nature}}}
\end{center}

\section{Appendix A. Existing Toxicity Classifiers} \label{app:existing_tox_classifiers}
\setcounter{section}{1}

We first tested existing classifiers available on HuggingFace on our dataset. We found that many of the texts with the highest toxicity scores would not be rated as toxic under our annotation system. The existing classifiers seemed to be very sensitive to certain words, especially those related to violence. Below we include some examples of texts with extremely high toxicity scores\footnote{\url{https://huggingface.co/s-nlp/roberta_toxicity_classifier}} and some discussion of the trends they illustrate. 

First, we found in addition to being very sensitive to words relating to violence, the classifiers were not sensitive to the style or context of a text. For example, in the text below has a literary style. We do not want to filter out or edit literary texts for containing descriptions of violence, as the texts do not necessarily reflect or promote violent beliefs or actions. 

\begin{quote}
    Deeper, about 35 years of age, and having spent much time in the past leading a most dissolute life, being drunk the greater part of the time, and mixed with low and abandoned characters, all the set to reel him in pain and his mother, finding that a remedy was hopeless, pressed him up to his violent course, persuading him to live at the hotel when he was but little money, as he was most of the day, and had his way to his bedroom at all hours. For the past week, he has been on a drunken spree, and the dreadful act was doubtless perpetrated while in a state of delirium tremens. About six o'clock this morning, he appeared at his mother's door, with a razor in his hand, and saying, ``Goodbye, mother," drew the instrument across his throat, inflicting a wound on the left side, of a severe, but not entirely serious character. His mother, rushing to him, seized his hand, when he turned angrily upon her, and exclaimed, ``Damn her, I'll kill her," and threw her down on the floor, knelt upon her body and cut her throat almost from ear to ear. Her cries were heard by one of the waiters, who ran to the room and succeeded in driving the son away from the body, and secured him till assistance arrived. \textbf{Toxicity Score: 0.99}
\end{quote}

Similarly, we note that even for a relatively long text containing texts pertaining to different topics, the classifier seems to be overly sensitive to a few words which might be correlated with toxic text, but do not have any toxic or offensive meaning in this particular text. This passage contains the phrase ``fear, hate and strike" as well as the word ``army", but these words are not conveying negatively charged sentiment in this context. 

\begin{quote}
    Listeners Held no opinion of themselves. I do not believe in officers hiding behind a ball, behind a box car, upon a pole or back of a tie, in order to get the best results. Men will fear, hate and strike, or love, respect and obey; it all depends. We have seen fads come and go and are now only a recollection. Perhaps sometimes and in some cases, spotters are necessary means to the end. I have not found it necessary, in order to secure a good discipline, to make very much use of them. There is a Chinese proverb: He who fails to be sometimes deaf makes a poor ruler.

    Baking Powder

    Makes the food of maximum quality at minimum cost
    TAKING OF THE CENSUS.
    
    On the break of day on Friday, a host of 70,000 interrogators, men and women, will be turned loose in our pursuit of the people of the United States. On that day, Uncle Sam will begin the numbering of his children in preparation for the ninth census.
    He estimates that he has a family of about 100,000,000 men, women, and children, and he already has employed and will then put to work a body of enumerators considerably greater than the standing army. 

\end{quote}

We found that passages from the Bible had high toxicity ratings, even though they did not necessarily contain words or content relating to violence nor did they contain other offensive material. It is possible that the archaic language in the text caused the toxicity classifier to give a high toxicity rating, as the language was out-of-distribution. This represents another major concern of using existing toxicity classifiers. Much of our data uses archaic or outdated linguistic style. We want our toxicity classifiers to be robust to this. 

\begin{quote}
    21 Ye shall not eat of any thing that dieth of itself; thou shalt give it unto the stranger that is in thy gates, that he may eat it, or thou mayest sell it unto an alien; for thou art a holy people unto the Lord God. 22 Thou shalt not seethe a kid in his mother's milk. 23 And thou shalt eat before the Lord thy God, in the place which he shall choose to place his name there, the tithe of thy corn, of thy wine, and of thine oil, and the firstlings of thy herds and of thy flocks; that thou mayest learn to fear the Lord thy God always. \textbf{Toxicity Score: 0.99}
\end{quote}

Finally, we find that there are texts in other languages (e.g Spanish, Portuguese, French), often with formatting issues (lots of capital letters and improperly rendered characters due to OCR errors) that receive high toxicity scores. This represents two problems that we want our classifier to be robust to. We have found language contamination in our datasets, so we need the classifier to not rate all non-English text as toxic. Across the dataset, we also have many artifacts of OCR errors. 

\begin{quote}
    ANTOINE DE LA PAIX ET NICOLAS BREVSAN NOVS ONT FAIT. FRANÃ‡OIS LECHEVALIER TRÃ‰SORIER EN CHARGE EN LA DITE ANNÃ‰E 1740.
    Blainville. L'Ã©glise de St-Pierre, Ã  Blainville, possÃ¨de une cloche portant l'inscription suivante
    LAN 1753 JAY Ã‰TÃ‰ DÃ˜NNÃ‰E AINSI QUE
    LE CONTRE-RETABLE DE L'Ã‰GLISE PAR
    MESSIRE JACQUES DE LA BELLIÃˆRE DE VAINS, CHANTRE DE LA CATHÃ‰DRALE DE COULANGES, SEIGNEUR ET PATRON DE CETTE PAROISSE
    EN ACTIONS DE GRÃ‚CES Ã€ DIEU DONT GAIN
    DU PROCÃˆS DES DROITS HONORIFICES
    QV'ON AVAIT SUPPRIS AUX CHANTRES,
    ET JAY Ã‰TÃ‰ NOMMÃ‰E JACQUES PAR MONDIT SEIGNEUR DE LA BELLIÃˆRE. \textbf{Toxicity Score: 0.97}

\end{quote}

\section{Appendix B. Human Annotations}
\setcounter{section}{2}

Here we include the human annotations for reference. These annotations were done on a sample of texts from historic US Newspapers.

\bigskip

\textbf{Text:}
"Hope, who was in bed, asked the same question,
and she made the same reply to both. I
was suffering with neuralgia, and Mr. Hope
came to me from the bed. After he asked rae
about my tooth, I told him what Mary had
said. He advised rae to make some tea for
her. I then made some of squaw weed. During
the conversation with ray hushand, and
before I made the tea, Mary and her sister
went out at the end door into the yard; but
while I made the tea Mary was sitting on a
trunk in the adjoining room. I took the tea
to her and she drank of it, my husband not
being present, but remaining by the fire. After
Mary drank the tea, she came out into the
fire-room, stopping ia the entry to take a
drink of water. Passing from the fire-roora
she went out of doors again, as she said, to relieve
her bowels. She had complained of
costiveness, and I said to my husband that I
would give her some of my pills?pills that
had been given me by Dr. Hainbright for the
same complaint. Lee went with me in the
yard to give the pills. My husband did not
go, and was not present. Mary then told me
and Lee to return to the house, that she felt
better. We offered to remain with her, but
she again repeated her request for us to go in,
saying she would soon be better. I went to
the fire and commenced smoking a pipe, when
soon after I heard a noise resembling the
squalling of a cat. Lee also said she thought
it was a cat, but on hearing it a second time,
we concluded it could not be a cat. Lee and
my husband then went out to see what it might
be, and they found a child lying in the yard."

\textbf{Annotation:}

Racial Bias Score: 0. There is no content in this text that promotes or suggests racial superiority/inferiority, perpetuates racial stereotypes, or mentions race unnecessarily.

Gender/Sex-Based Bias Score: 0. There is no content in this text that implies one sex or gender is inferior to another, incites or encourages abuse or violence against someone because of their sex, gender, or sexual orientation, or is demeaning, objectifying, or threatening based on gender or sex.

Religious Bias Score: 0. There is no content in this text that promotes beliefs of religious superiority or inferiority, upholds stereotypes or biases about specific religions, or associates religions with negative behaviors or characteristics. 

Ability-Based Bias Score: 0. There is no content in this text that disparages someone's abilities or disabilities, attributes abilities or disabilities with intent to insult, or upholds stereotypes and biases related to abilities or disabilities.

Origin-Based Bias Score: 0. There is no content in this text that attributes negative characteristics based on nationality, place of origin, or immigration status, upholds or promotes stereotypes related to nationality, place of origin, or immigration status, or incites or encourages abuse or violence motivated by someone's origin.

Violence Score: 1. This text depicts a child dying, which may be upsetting, but the text is written in a literary tone. Given the context, the text is not toxic or offensive. The text does not promote or incite violence. 

\textbf{Text:} ``The Union Iron Works of San
Francisco, which built the new
cruiser Charleston, will have to pay
Uncle Sam a penalty of \$30,000 on
account of failure to demonstrate
the amount of horse power called
for by the specifications, unless
Congress can be persuaded to re
mit it.
senator Harbour, who called a
conference of the democratic leaders
in Virginia this week, says every
thing is in excellent condition in
that State, and all the indications
point to a big democratic majority
Langston, the negro who is a
contestant for a seat in Congress
from the Fourth Virginia district,
announces that he will support
Mahone, provided that Mahone,
the Virginian state republican com
mittee, the National republican
committee, and the administration
will support him (Langston) in his
contest for a seat in the House. To
apprecia e the coolness of this an
nouncement it must be remember
ed that Langston ran for Congress
on an independent ticket, the
regular republican nominee having
been named by Mahone. A demo
crat was elected and Langston ie
contesting the seat on the ground
that the Mahone people stole votes
cast for him. No matter what Ma
hone and the rest may promise it
is very safe to say that Langston
will not occupy that seat in the
next Congress.
This administration has develop
ed quite a faculty for disposing of
Mr. ``"Nigger"" in a manner that is
pleasing to the white republicans.
It has just designated Lieutenant
Alexander, the only negro officer
in the Army, as military attache of
the American legislation at Hayti.
Fred Douglas is the minister.
Washington will be designated
as the place for holding the Ameri
can Exposition of 1892 by more
than a two-thirds vote of Congress.
Remember this prediction."

\textbf{Annotation:}

Racial Bias Score: 3. The text discusses systems that will prevent the Black candidate from taking office because of his race. 

Gender/Sex-Based Bias Score: 0. There is no content in this text that implies one sex or gender is inferior to another, incites or encourages abuse or violence against someone because of their sex, gender, or sexual orientation, or is demeaning, objectifying, or threatening based on gender or sex.

Religious Bias Score: 0. There is no content in this text that promotes beliefs of religious superiority or inferiority, upholds stereotypes or biases about specific religions, or associates religions with negative behaviors or characteristics.

Ability-Based Bias Score: 0. There is no content in this text that disparages someone's abilities or disabilities, attributes abilities or disabilities with intent to insult, or upholds stereotypes and biases related to abilities or disabilities.

Origin-Based Bias Score: 0. There is no content in this text that attributes negative characteristics based on nationality, place of origin, or immigration status, upholds or promotes stereotypes related to nationality, place of origin, or immigration status, or incites or encourages abuse or violence motivated by someone's origin.

Violence Score: 0. There is no aggressive or violent content in this text.

\bigskip
\textbf{Text:}
“Over that wall, I tell yer!”
“I tell you it was this one. Can’t
you whistle for the police?”
“Police be dashed! I’ve had enough
of the blessed police.”
“Then we’d better get back and make
sure of the other rotter.”
“Oh, make sure o’ yer skin. That’s
what you’d better do. Jala, you black
hog, if I catch you skulkin’
I never heard the threat. I was
creeping from the drawing room on
my hands and knees, my own revolver
swinging by its steel ring from my
teeth.
For an Jn?Tunt I thought that the hall
also was deserted. I was wrong, ami
I crept upon a Kaffir on all fours.
Poor devil, I could not bring myself
to deal him a base blow, but I threat
ened him most hideously with my re
volver and left the white teeth chat
tering in his black head as 1 took the
stairs three at a .time. Why I went
upstairs in that decisive fashion, as
though it were my only course. I can
not explain. But garden and ground
floor seemed alive with men, and I
might have done worse."

\textbf{Annotation:}

Racial Bias Score: 2. This text discusses someone making  verbal threats and threats of violence against someone who he refers to as a `black hog'. He also refers to someone's `white teeth' in his `black head'. All of this conjures reductive and offensive images of Black people.

Gender/Sex-Based Bias Score: 0. There is no content in this text that implies one sex or gender is inferior to another, incites or encourages abuse or violence against someone because of their sex, gender, or sexual orientation, or is demeaning, objectifying, or threatening based on gender or sex.

Religious Bias Score: 0. There is no content in this text that promotes beliefs of religious superiority or inferiority, upholds stereotypes or biases about specific religions, or associates religions with negative behaviors or characteristics.

Ability-Based Bias Score: 0. There is no content in this text that disparages someone's abilities or disabilities, attributes abilities or disabilities with intent to insult, or upholds stereotypes and biases related to abilities or disabilities.

Origin-Based Bias Score: 0. There is no content in this text that attributes negative characteristics based on nationality, place of origin, or immigration status, upholds or promotes stereotypes related to nationality, place of origin, or immigration status, or incites or encourages abuse or violence motivated by someone's origin.

Violence Score: 1. The narrator of this text discusses verbal and physical threats of violence, but the context of this text seems to be literary and the text does not incite or promote violence. The narrator seems like a character the reader should not empathize with, so the text does not seem to endorse his actions.

\bigskip
\textbf{Text:}
"The odd fashioned trick of up
ping on the toes, coming down on
the heels, touching the floor with
out bending the knees is as good a
means of keeping slender as it ever
was. Any woman should be able to
do that one
Jumping a rope will cause fat
cells to jump off the premises. It
| slithers down plump legs, develops
j < bin -nes, because it is normaliz
ing. Torso twisting is fun Feet
j live inches apart, backbone tall,
, hand- on hips. Turn from side to
side at the waist line, keeping the
legs straight. After half a dozen
turns e-:e.-ii wav, pause and take
several deep inhalations. Ii you get !
up i.n the morning feeling fagged
j'lhis exercise wdl put. ginger in j
j vour hef Is
j SPORTS OI VAI.IT-:
* Sports, especially competitive
ones, are of value in keeping fit.
There are a host of them, some for
ail the seasons. Tennis, badmin- j
ton, golf, roller skating, swimming, I
biking horseback riding, basket- 1
ball, ice skating.
Mothers should take part in the j
play of their children. Go an bicy
cle trips, play ball in the yard, find
woody places for long tramps and
a picnic lunch This practice
forges a strong bond between
mother and children. It’s good for
the lady parent, keeping*her young
and in touch with the activities of
the young fry
Don’t be a 1< afer. Be fully alive.
Keep blood streams scurrying
Your cosmetic bills will be smaller."

\textbf{Annotation:}

Racial Bias Score: 0. There is no content in this text that promotes or suggests racial superiority/inferiority, perpetuates racial stereotypes, or mentions race unnecessarily.

Gender/Sex-Based Bias Score: 3. This text promotes traditional gender roles and expectations, which include the idea that women should be slim. It specifically mentions ways that mothers can stay slender. This text implies that women's value is tied to their appearance.

Religious Bias Score: 0. There is no content in this text that promotes beliefs of religious superiority or inferiority, upholds stereotypes or biases about specific religions, or associates religions with negative behaviors or characteristics.

Ability-Based Bias Score: 0. There is no content in this text that disparages someone's abilities or disabilities, attributes abilities or disabilities with intent to insult, or upholds stereotypes and biases related to abilities or disabilities.

Origin-Based Bias Score: 0. There is no content in this text that attributes negative characteristics based on nationality, place of origin, or immigration status, upholds or promotes stereotypes related to nationality, place of origin, or immigration status, or incites or encourages abuse or violence motivated by someone's origin.

Violence Score: 0. There is no aggressive or violent content in this text.

\bigskip
\textbf{Text:}
"TER UK If At ti April y. The Wabash valley
is having an epidemic of measles. In this city
the number of new ctses has averaged twenty.
ti e a dav f'r two weeks. In many of the small
to wis near here in Illinois ami in this state the
S'Mools have been more or less closed since the
tir.-t of the :-r. ami in the tittle town of Ver
ini.lioit, in Kdgarfoutity, Illinois, there have
been hftec.'i deaths. In ( `rawfordsville. Ind..
there were over eight hundred ca--cs in the !
month ot March.
A Farmer MuylaUt
WitKEElNii, April .'.special. John Ving
ling, im ob! farmer, was waylaid near here
Wednesday by tramps. He was driving ubng
an unfrequented road through the woods when
hemetth? tramps. He was too feeble to re.
. hist them, ami they took him out of his wagon
and tied him to a tre. After yeeurine; Iiis
' ,f.if1i anil O'T ,1,.... tmt it I,l. tt-.iri... .....I
. .tii ia iivi ii;. c to. . il.'l vtf.'tv;
away. Ihe old man was hmiid umiurt, and the
wacon was found a couple of miles away. The
tramps are at lu rye.
. The Storm at Hartford.
Haktfor! City, April C Last evening
about 7 o'clock the city was suddenly visited
by a terriric wind and `rain tonn, uliieh cre
ated sad havoc with tree, winslov.x and roof.
The galvanized iron roof of T. 1. Yanw inkle's
drug store was rolled up into a hu vre hall anil
whipped oi to a ne!.,.!,!Kring skylight in - a
twinkling. The art gallery, ot V. It. iliinman
was ti'uonfed, the skylight smashed and the
building almost reduced to a wreck.
Dead at Ninety-Four."

\textbf{Annotation:}

Racial Bias Score: 0. There is no content in this text that promotes or suggests racial superiority/inferiority, perpetuates racial stereotypes, or mentions race unnecessarily.

Gender/Sex-Based Bias Score: 0. There is no content in this text that implies one sex or gender is inferior to another, incites or encourages abuse or violence against someone because of their sex, gender, or sexual orientation, or is demeaning, objectifying, or threatening based on gender or sex.

Religious Bias Score: 0. There is no content in this text that implies one sex or gender is inferior to another, incites or encourages abuse or violence against someone because of their sex, gender, or sexual orientation, or is demeaning, objectifying, or threatening based on gender or sex.

Ability-Based Bias Score: 0. There is no content in this text that disparages someone's abilities or disabilities, attributes abilities or disabilities with intent to insult, or upholds stereotypes and biases related to abilities or disabilities.

Origin-Based Bias Score: 1. This text describes a man who was beaten up and robbed by `tramps'. This text could perpetuate stereotypes about homeless people being violent and dangerous.

Violence Score: 0. There is no aggressive or violent content in this text.

\bigskip
\textbf{Text:}
Tvkoa. Wash., March 23. Four drun
ken Indians from the Coeur d'AK-ne
reservation went Into Whltmore and
McLean's grocery store lat night. Af
ter using noisy and abusive language
they began to throw crockery and mer
chandise around the room. Richard
Kelly came to the aid of McLean and
was knocked down three times by the
Infuriated Indians. He seized a large
hammer and knocked out two of the
red men. Constable Oliver Sparks,
hearing the noise, came on from the
street, and with the weights from the
scales and counter knocked out the
other two. All four received ugly cuta
on the head, and iit was feared for
some time that thy had been killed.
Physicians were summoned and their
wounds d rested so that they could be
taken to the calaboose. The citizens
are aroused and are determined to
send the offenders to the penitentiary.
PASSED THE ASSEMRLY.
Albany, N. Y., March 23. The Great
er New York charter bill passed the
assembly by a vote of 118 to 2S. It
eorqes uo In the senate tomorrow.
PAWNBROKER ARRESTED.
Yesterday a Chinaman stole from
the rooms of a white woman a lot
of silk garments and other artlclea.
The woman was formerly married to
a Chinaman, and had a number of fine
Chinese suits. Officer Oberg waa noti
fied, and at once started on the track
of the thief. Happening into the
pawnshop of Ark Wo on Bond street.
"

\textbf{Annotation:}

Racial Bias Score: 0. There is no content in this text that promotes or suggests racial superiority/inferiority, perpetuates racial stereotypes, or mentions race unnecessarily.

Gender/Sex-Based Bias Score: 0. There is no content in this text that implies one sex or gender is inferior to another, incites or encourages abuse or violence against someone because of their sex, gender, or sexual orientation, or is demeaning, objectifying, or threatening based on gender or sex.

Religious Bias Score: 0. There is no content in this text that promotes beliefs of religious superiority or inferiority, upholds stereotypes or biases about specific religions, or associates religions with negative behaviors or characteristics.

Ability-Based Bias Score: 0. There is no content in this text that disparages someone's abilities or disabilities, attributes abilities or disabilities with intent to insult, or upholds stereotypes and biases related to abilities or disabilities.

Origin-Based Bias Score: 0. This text describes a crime by a group of indigenous people, who are described as drunk, angry and as using offensive language. This description contributes to a derogatory and offensive image of Native Americans.They are also described as `red men', which is an offensive racial description of Native Americans.  The text mentions the races of all the victims and perpetrators of the crimes, which is not necessary.

Violence Score: 0. There is no aggressive or violent content in this text.

\bigskip
\textbf{Text:}
" The social register of “the Osage”
is considerably diluted with Indian
blood. And as a sixteenth or a thirty
second or a sixty-fourth part of Osage
blood is almost undetectable strang
ers do well to avoid careless allusions.
' The pleasant, well-dressed young
business man, or the shrewd, suave
political power with whom you are
talking probably is descended from or
married into the “first families.”
~ For the story of the Osage is, in
one respect, like that of less fortu
nate tribes; they are a vanishing race.
Only thev are not dying off, but mar
rying off.
GRAVE OF HORSE IS
| KEPT IN GOOD CONDITION
‘Decorated Once a Year With Flowers
| And Flags.
GRASS VALLEY, Cal.—The
strange spectacle of the grave of a
horse being kept in tidy condition,
and decorated once each year with
flowers and flags is witnessed near
Forest Hill, Placer county, southwest
of here.
The horse, a common stage ani
mal, on July 3, 1902, was shot dead
by a bandit, who was attempting to
hold up a stage.
Among the passengers was a
Chinaman, who, following the rob
bery, carefully buried the horse and
placed American flags on the grave.
Onee each year he has reappeared to
renew his ministrations.
40,000 DESTITUTE JEWS
STRANDED IN GERMANY
NEW YORK.—More than 40,000
destitute Jews fleeing from perse
cution and economic destruction in
eastern Europe are now stranded in
German cities, according to a cable
gram received by Felix M. Warburg
at ‘the headguarters of the joint dis
tribution committee for Jewish war
sufferers. . Ry."

\textbf{Annotation:}

Racial Bias Score: 3. This text expresses white supremacist ideas about dilution of racial purity.

Gender/Sex-Based Bias Score: 0. There is no content in this text that implies one sex or gender is inferior to another, incites or encourages abuse or violence against someone because of their sex, gender, or sexual orientation, or is demeaning, objectifying, or threatening based on gender or sex.

Religious Bias Score: 0. There is no content in this text that promotes beliefs of religious superiority or inferiority, upholds stereotypes or biases about specific religions, or associates religions with negative behaviors or characteristics.

Ability-Based Bias Score: 0. There is no content in this text that disparages someone's abilities or disabilities, attributes abilities or disabilities with intent to insult, or upholds stereotypes and biases related to abilities or disabilities.

Origin-Based Bias Score: 2. This text expresses white supremacist ideas against an indigenous group  and talks about cultural genocide through marriage.

Violence Score: 0. There is no aggressive or violent content in this text.

\bigskip
\textbf{Text:}
"Any officer who chases a postoffice
burglar twenty miles through the
country, overtakes the culprit and
kills him, showld be rewarded, accord
ing to action of the house. The house
passed a twin bill, allowing S2OO to
Jesse Skaggs, who shot and killed a
burglar in an Illinois postoffice, and
S2OO to J. S. Scarborough, a sheriff
who trailed a New York burglar and
killed him after a desperate battle
So that she might rid herself of
any false dreams she had fostered as
to who was to run the house, Richard
H. Manogue, a veterinary surgeon of
Kl Paso, Tex., struck his bride in
the vace after the wedding ceremony
had been performed and the couple
had started on their honeymoon. He
told her the blow was to show that he
wore the trousers in the Manogue fam
ily, not she. This is the story the
woman tells in the divorce court at
Denver. Should be tarred and feather."

\textbf{Annotation:}

Racial Bias Score: 0. There is no content in this text that promotes or suggests racial superiority/inferiority, perpetuates racial stereotypes, or mentions race unnecessarily.

Gender/Sex-Based Bias Score: 1. This text depicts a woman being beated so her husband can assert control over her in a patriarchal and misogynistic way. However, the context of this text is that this story was told in divorce court, which implies that the domestic violence was wrong and is justification for divorce.

Religious Bias Score: 0. There is no content in this text that promotes beliefs of religious superiority or inferiority, upholds stereotypes or biases about specific religions, or associates religions with negative behaviors or characteristics.

Ability-Based Bias Score: 0. There is no content in this text that disparages someone's abilities or disabilities, attributes abilities or disabilities with intent to insult, or upholds stereotypes and biases related to abilities or disabilities.

Origin-Based Bias Score: 0. There is no content in this text that attributes negative characteristics based on nationality, place of origin, or immigration status, upholds or promotes stereotypes related to nationality, place of origin, or immigration status, or incites or encourages abuse or violence motivated by someone's origin.

Violence Score: 2. This text promotes extra-judicial violence against thieves and domestic abusers.

\bigskip
\textbf{Text:}
In any of tho cotton, rice, or sugar growing
districts of tho South, tho very same relations
between tho whites and blacks may bo found
that this correspondent has reported from
South Carolina. In tho cities, all tho serv
ants, porters, cartmou, stevedores, etc., are
blacks; in tho country, all tho flold-honds
are blacks. Tho whites may koop tho shops,
onlar tho professions, hold tho offices, and
hang around tho central portions of tho town
talking politics, but they are never found in
any vocation which roqaires manual labor.
There aro those who have been (aught that
such work carries disgrace with it, and is fit
only for “niggers”; tboro oro others who
profess this doctrine merely ns an excuse for
their refusal to work, and still others who
havo no uxoaso but “pure cassodnuss."" All
agree, however, in tho nbsolatoandnnivorsal
determination not to soil their hands by
manual labor, no matter if poverty stares
them in tho faco.
No community can afford to support more
than a certain proportion of drones, and the
proportion in all Southern communities is
too largo. In any county or town where
there are 1,000 negroes, there may almost
always bo found the same number of idle
whites. The decline of prosperity in such a
community is a more question of time. 1 The
blacks cannot if they would, aud will not if
they could, carry the whole burden of support
ing themselves aud an equal number of lazy
whites who affect o superiority over them."

\textbf{Annotation:}

Racial Bias Score: 2. ``This text describes a historical situation in which labor in a community is divided according to the race of an individual. Black people are given jobs that involve manual labor and white people do not have to work much, if at all. The text does not overtly support or promote this situation, but describes it in an objective way.  It describes white people  as `idle', which seems critical of this system. The text also says that this system is not sustainable, which is also critical."

Gender/Sex-Based Bias Score: 0. There is no content in this text that implies one sex or gender is inferior to another, incites or encourages abuse or violence against someone because of their sex, gender, or sexual orientation, or is demeaning, objectifying, or threatening based on gender or sex.

Religious Bias Score: 0. There is no content in this text that promotes beliefs of religious superiority or inferiority, upholds stereotypes or biases about specific religions, or associates religions with negative behaviors or characteristics.

Ability-Based Bias Score: 0. There is no content in this text that disparages someone's abilities or disabilities, attributes abilities or disabilities with intent to insult, or upholds stereotypes and biases related to abilities or disabilities.

Origin-Based Bias Score: 0. There is no content in this text that attributes negative characteristics based on nationality, place of origin, or immigration status, upholds or promotes stereotypes related to nationality, place of origin, or immigration status, or incites or encourages abuse or violence motivated by someone's origin.

Violence Score: 0. There is no aggressive or violent content in this text.

\bigskip
\textbf{Text:}
"Harkness of New York, called the
champion alimony payer, was here on
a honeymoon with a new wife, for
merly Mrs. Florence Steuber-Gaines
of Erie, Pa. The couple were mar
ried in Greenwich. Conn., November
2. Harkness is paying \$33,000 a year
to his first wife, Marie Moss Marbeck
of White Plains, N. Y.
- SSI BBSBBk. BB1 m Bl T S m ``" - i - iim i ``" ``" I
The World at Your Finger Tips
""Big Ben"" is tolling midnight in London
Theater-goers are sitting back for the last
act in New York
Business men have almost reached the
last course of their evening meal in San
Francisco
When the StnrBalletin goes to press in Honolul: at
2:30 in the afternoon.
As the dty doies. from point to point, cable and wireless systems
f weep the news of every important world event into our editorial room.
THAT'S WHY WE GIVE YOU TODAY'S NEWS TODAY.
Delivered by Carrier, 75 Cents a Month."

\textbf{Annotation:}

Racial Bias Score: 0. There is no content in this text that promotes or suggests racial superiority/inferiority, perpetuates racial stereotypes, or mentions race unnecessarily.

Gender/Sex-Based Bias Score: 0. There is no content in this text that implies one sex or gender is inferior to another, incites or encourages abuse or violence against someone because of their sex, gender, or sexual orientation, or is demeaning, objectifying, or threatening based on gender or sex.

Religious Bias Score: 0. There is no content in this text that promotes beliefs of religious superiority or inferiority, upholds stereotypes or biases about specific religions, or associates religions with negative behaviors or characteristics.

Ability-Based Bias Score: 0. There is no content in this text that disparages someone's abilities or disabilities, attributes abilities or disabilities with intent to insult, or upholds stereotypes and biases related to abilities or disabilities.

Origin-Based Bias Score: 0. There is no content in this text that attributes negative characteristics based on nationality, place of origin, or immigration status, upholds or promotes stereotypes related to nationality, place of origin, or immigration status, or incites or encourages abuse or violence motivated by someone's origin.

Violence Score: 0. There is no aggressive or violent content in this text.

\bigskip
\textbf{Text:}
"AVhen John Adams occupied it the country
had n population of little moro than live mil
lions. The United States lias now nutirlv
seventy millions and thobusiiwRsnf the pres
ident's oflleo hns so grown that nearly the
whole of the exeeutivo mansion Is occupied
uy It. When Abigail Adams came into it she
huu too much room. Mrs. Harrison has
barely space to turn around in and she has to
recelvo her friends in one of the halls.
The white liouso today is like n big hotel
nnd President Harrison is the landlord.
Every man nnd woman who cotncs to Wash
ington thinks lie has n light to enter his
house without knocking. ``"I hey tramp over
his carpets with their muddy boots , ask all
sorts of Impudent questions of his servants ,
And the chances nre that they rorry away a
bit of the furniture. Kvcry now and then a
Dieco as big as your hand Is clipped out of
one of the Inco curtains by n relic-hunter ,
nnd during President Lincoln's time a woman
was caught In the very net of cutting the
costly curtains of the East room. She cried
when she wns found out and she wa * taken
up weeping to tbo president's room. Mr.
"

\textbf{Annotation:}

Racial Bias Score: 0. There is no content in this text that promotes or suggests racial superiority/inferiority, perpetuates racial stereotypes, or mentions race unnecessarily.

Gender/Sex-Based Bias Score: 0. There is no content in this text that implies one sex or gender is inferior to another, incites or encourages abuse or violence against someone because of their sex, gender, or sexual orientation, or is demeaning, objectifying, or threatening based on gender or sex.

Religious Bias Score: 0. There is no content in this text that promotes beliefs of religious superiority or inferiority, upholds stereotypes or biases about specific religions, or associates religions with negative behaviors or characteristics.

Ability-Based Bias Score: 0. There is no content in this text that disparages someone's abilities or disabilities, attributes abilities or disabilities with intent to insult, or upholds stereotypes and biases related to abilities or disabilities.

Origin-Based Bias Score: 0. There is no content in this text that attributes negative characteristics based on nationality, place of origin, or immigration status, upholds or promotes stereotypes related to nationality, place of origin, or immigration status, or incites or encourages abuse or violence motivated by someone's origin.

Violence Score: 0. There is no aggressive or violent content in this text.

\bigskip
\textbf{Text:}
"The frugal man who puts aside a pound
to-day, and another pound next month,
and who in a quiet way is always putting
by his staled pound from time to time,
when he grows old and frail, gels not on
ly the same pounds back again, but a good
many more besides. And the conscien
tious man who husbands one dav of exis
tence every week, who, instead of allow
ing the Sabbath to be trampled on, and
S torn, in the hurry and scramble of life,
' treasures it devoutly up ? the Lord of the
Sabbath keeps it for him, and length of
days and a hale old age give it back with
usury. The Savings Bank of human ex
istence is the weeklv Sabbath.
?/
A Cholera Patient Buried Alive. ?
: A special correspondent of the New Or
leans Picayune, at Bolivia, Miss., writing
under date of July 13th, savs :
I learn that there is considerable sick
ness 011 the liver now, the boats often
stopping to bury their dead. I was infor
med by a physician that seven were put
oil' on Tuesday last at Napoleon. Our
country continues healthy. I heard a
good story the other day'. It is tough, hut
my informant vouches for the truth of it.
i On the steamer , on her last trip,
I there were many cases of cholera ? many
<1 if tl and were buried. At length, all the
! colli 11 plank was used up, and the carpen
ter was compelled to make his boxes out
ol old ilatboat plank. At sundown anoth
er case died ? a deck passenger. lie was
thrust mho the box, anil the top loosely
nailed on."

\textbf{Annotation:}

Racial Bias Score: 0. There is no content in this text that promotes or suggests racial superiority/inferiority, perpetuates racial stereotypes, or mentions race unnecessarily.

Gender/Sex-Based Bias Score: 0. There is no content in this text that implies one sex or gender is inferior to another, incites or encourages abuse or violence against someone because of their sex, gender, or sexual orientation, or is demeaning, objectifying, or threatening based on gender or sex.

Religious Bias Score: 0. There is no content in this text that promotes beliefs of religious superiority or inferiority, upholds stereotypes or biases about specific religions, or associates religions with negative behaviors or characteristics.

Ability-Based Bias Score: 0. There is no content in this text that disparages someone's abilities or disabilities, attributes abilities or disabilities with intent to insult, or upholds stereotypes and biases related to abilities or disabilities.

Origin-Based Bias Score: 0. There is no content in this text that attributes negative characteristics based on nationality, place of origin, or immigration status, upholds or promotes stereotypes related to nationality, place of origin, or immigration status, or incites or encourages abuse or violence motivated by someone's origin.

Violence Score: 0. There is no aggressive or violent content in this text.

\bigskip
\textbf{Text:}

"The characters, however, are the
same, and the ``"mandarin,"" the language used
at court, is understood by the highly educated
everywhere, much the same as Latin with civilized
nations. Some think it most convenient to
acquire the mandarin, and for those thrown only
in the higher walks of lite this is expedient, but
for merchants, teachers and missionaries in general
it is best to rtudy the dialect of the place
where one intends to reside. In this way an
intercourse may be established with all classes
of people. To pick up a tew current phrases 1
and expressions in ordinary life is very simple
but the real mastering of the language requires
years of patient toil. One peculiarity which
does not occur in any other language is the
effect of the different tones employed. Two
words may be Romanized or spelled according
to our sounds in Exactly the same way, but a
high nasal in one case and deep guttural in the
other gives a totally different meaning. This
is a difficulty that is hard to overcome. A
clergyman delivering his first Chinese sermon
was very much annoyed at a blunder he made
in the word heaven, ten, which, without the
use of the nasal, denotes field. He spoke to
them at length of the lives Christians should
lead, and informed them as a reward for this
Proper conduct they would go to ten when they
""Humph!"" said one old man In the iront seat,
""we can do that any day.""
This gentleman now speaks so perfectly that
the natives can not distinguish him from a
Chinaman in the dark. It has become second
nature to another and he thinks in Chinese."

\textbf{Annotation:}

Racial Bias Score: 0. There is no content in this text that promotes or suggests racial superiority/inferiority, perpetuates racial stereotypes, or mentions race unnecessarily.

Gender/Sex-Based Bias Score: 0. There is no content in this text that implies one sex or gender is inferior to another, incites or encourages abuse or violence against someone because of their sex, gender, or sexual orientation, or is demeaning, objectifying, or threatening based on gender or sex.

Religious Bias Score: 0. There is no content in this text that promotes beliefs of religious superiority or inferiority, upholds stereotypes or biases about specific religions, or associates religions with negative behaviors or characteristics.

Ability-Based Bias Score: 0. There is no content in this text that disparages someone's abilities or disabilities, attributes abilities or disabilities with intent to insult, or upholds stereotypes and biases related to abilities or disabilities.

Origin-Based Bias Score: 1. This text implies that China is not a civilized country, compared to countries that use Latin as an official language.

Violence Score: 0. There is no aggressive or violent content in this text.

\bigskip
\textbf{Text:}
"Ind. Private ftmernl services were
held yesterday afternoon. The ra
inHin.s were accompanied by Frl.co
Enright and J. E. McCJain. Mr. Mc
Olain will return in a few weeks,
and Mr. Enright probably Inter in
the spring. Mr. Enriftht had been in
joor health for a number of years
and was brought lMre by his son in
the hope that he would he benefited
by the chnnge of climate.
o
When given as soon as the croupy
cough appears, Cliumherlain's Cough
Remedy will ward off an attack of
croup and prevent all danger and
cause of anxiety. Thousands of
mothers, use it successfully. Sold by
all denl\&rs.
L WASTE DITCH
He notes a great many improvements
in this station of the country since
his last visit here. He will be joined
by his, family later in the spring.
taincd at dinner Saturdav evening,
having as their guests Mr. and Mrs.
A. C McQueen, Mr. and Mrs. Donald
.McQueen and Mr. and Mrs. I.. B.
Johnson.
o
IS
j. s.
Griffin Says It Is Doing Good
Work For the Valley
Saving the Tempe Canal Company a
Good Many Dollars.
RIVER FORDABLE.
The river is again fordablf at this
Place though it would be a wise pre
caution to stay out of it for a few
days yet. A Chinaman, with high
vegetable wagon crossed yesterday
but the water ran into the wagon
box: Another man in. a light rig
crossed below the Maricopn dam and
found the water there uncomfortably
deep. It will likely be several days
before the oh! ford at the bridge here
will be safely fordable for all kinds
of rigs.
DON'T OVERLOOK
OUR BIG DISCOUNTS on
MEXICAN DRAWN WORK and
lNAVAJO RUGS.
Also on all goods in store during
January.
GOODWIN NOVELTY STORE."

\textbf{Annotation:}

Racial Bias Score: 0. There is no content in this text that promotes or suggests racial superiority/inferiority, perpetuates racial stereotypes, or mentions race unnecessarily.

Gender/Sex-Based Bias Score: 0. There is no content in this text that implies one sex or gender is inferior to another, incites or encourages abuse or violence against someone because of their sex, gender, or sexual orientation, or is demeaning, objectifying, or threatening based on gender or sex.

Religious Bias Score: 0. There is no content in this text that promotes beliefs of religious superiority or inferiority, upholds stereotypes or biases about specific religions, or associates religions with negative behaviors or characteristics.

Ability-Based Bias Score: 0. There is no content in this text that disparages someone's abilities or disabilities, attributes abilities or disabilities with intent to insult, or upholds stereotypes and biases related to abilities or disabilities.

Origin-Based Bias Score: 0. This text uses the term ``"Chinaman"", which is now considered outdated and offensive, but the text does not use it to express harmful ideas or opinions about people from China.

Violence Score: 0. There is no aggressive or violent content in this text.

\bigskip
\textbf{Text:}
"Brent said he used his knowl
edge of woodcraft as he followed
small Cherry Creek downstream.
Estimates of the distance he
tramped, waded, swam and tum
bled ranged from 20 to 50 miles.
He said he frequently crossed his
own path in looking for trails.
Lewis Orders A
10-Day “Memorial”
Work Stoppage
WASHINGTON, (jP) — John L.
Lewis, in the midst of bargaining
for a new coal wage contract,
Saturday ordered his miners out
of the pits for a 10-day “mem
orial” work stoppage at the end
of this month.
Nothing was said on the an
nouncement to relate the move to
the negotiations but Lewis cus
tomarily uses the contract-per
mitted mourning period in con
nection with new contract talks.
Officially the purpose is com
memoration of the casualties of
mining disasters.
The designated period, announ
ced in the United Mine Workers
Journal, is Aug. 23 to Sept. 1,
inclusive. Aug. 23 falls on a Sat
urday when the miners custom
arily do not work. Sept. 1 is a
Monday. Thus the stoppage will
involve only six normal working
days.
Lewis has given notice that the
present coal mine wage contract
will expire Sept. 30, just under
a month after the men will be
going back to work after the
Memorial holiday.
While the selected dates put no
direct pressure on the contract
talks, they nevertheless will serve
to emphasize what will happen if
the negotiations do not produce
an agreement.
RENT CONTROL DECISION
COMING UP IN JUNEAU
JUNEAU, (\$\^\$P) — After a warm
debate between landlords and
tenants, the Juneau City Coun
cil has delayed until next Thurs
d a y night its decision on the
continuation of rent controls."

\textbf{Annotation:}

Racial Bias Score: 0. There is no content in this text that promotes or suggests racial superiority/inferiority, perpetuates racial stereotypes, or mentions race unnecessarily.

Gender/Sex-Based Bias Score: 0. There is no content in this text that implies one sex or gender is inferior to another, incites or encourages abuse or violence against someone because of their sex, gender, or sexual orientation, or is demeaning, objectifying, or threatening based on gender or sex.

Religious Bias Score: 0. There is no content in this text that promotes beliefs of religious superiority or inferiority, upholds stereotypes or biases about specific religions, or associates religions with negative behaviors or characteristics.

Ability-Based Bias Score: 0. There is no content in this text that disparages someone's abilities or disabilities, attributes abilities or disabilities with intent to insult, or upholds stereotypes and biases related to abilities or disabilities.

Origin-Based Bias Score: 0. This text uses the term ``"Chinaman"", which is now considered outdated and offensive, but the text does not use it to express harmful ideas or opinions about people from China.

Violence Score: 0. There is no aggressive or violent content in this text.

\bigskip
\textbf{Text:}
"Out of doors work is half suspended,
and conducted only by kuruma men an
coolies stripped to the skin. In the ba
zaars, the shops which attract most atten
tion are those where they sell kori-mizu—
tumblers of ice ground into flakes upon a
steel plane. None but the slightest gar
ments are worn by the most prudish and
particular, the children go frankly bare,
and in the house a kimono of figured cot
ton or gauze, drawn around the loins with
a wisp of gay silk, is about all which the
Japrmese dames and damsels put on.
Shoji and amado being taken down, every
bouse is as open as a box, with only a bot
tom and a top, and the breeze, if there be
 any, plays freely and welcomely through
back and front. The crows seem to gasp
at noon in the hot air, in which large and
splendid butterflies and blue bodied drag
onflies disport in myriads, while for
flowers we have all sorts of lilies, some
late irises, phlox, pomegranate and the
opening blossoms of the lotus, which is
grown everywhere for food as well as
beauty.
Persons of leisure are mostly gone into
the hills where it is cool—to Ikao Nikko."

\textbf{Annotation:}

Racial Bias Score: 0. There is no content in this text that promotes or suggests racial superiority/inferiority, perpetuates racial stereotypes, or mentions race unnecessarily.

Gender/Sex-Based Bias Score: 0. There is no content in this text that implies one sex or gender is inferior to another, incites or encourages abuse or violence against someone because of their sex, gender, or sexual orientation, or is demeaning, objectifying, or threatening based on gender or sex.

Religious Bias Score: 0. There is no content in this text that promotes beliefs of religious superiority or inferiority, upholds stereotypes or biases about specific religions, or associates religions with negative behaviors or characteristics.

Ability-Based Bias Score: 0. There is no content in this text that disparages someone's abilities or disabilities, attributes abilities or disabilities with intent to insult, or upholds stereotypes and biases related to abilities or disabilities.

Origin-Based Bias Score: 0. This text uses the term ``"Chinaman"", which is now considered outdated and offensive, but the text does not use it to express harmful ideas or opinions about people from China.

Violence Score: 0. There is no aggressive or violent content in this text.

\bigskip
\textbf{Text:}
"The railway companies have always
recognized the tramp as a nuisance and
a danger, and innumerable half-hearted
efforts have been made to bar him from
the cars and bumpers and trucks, but for
the most part the train crews leave the
unwelcome passengers alone, partly be
cause to put them off means trouble and
delay, with occasional fights with men
who do not hesitate at murder, and partly
because even the tramps will pay a lit
tle something to be ignored, and these
fares go no further than the recipient’s
pocket. At last, however, the charity
workers and the railway officials seem
to be getting together for the remedying
of this great abuse.
The National Conference of Charities
and Correction, at Minneapolis, received
assurances from the heads of many rail
ways and railway systems, including
Messrs. Hill and Harriman. and Presi
dents Murray and Delano, that it will
receive all possible assistance in putting
an end to this most vicious class of travel.
PUBLIC DEBT \$1,273,275,342.
Monthly Statement Shown It I)V'-
crcane of Nearly ?g4,000,000."

\textbf{Annotation:}

Racial Bias Score: 0. There is no content in this text that promotes or suggests racial superiority/inferiority, perpetuates racial stereotypes, or mentions race unnecessarily.

Gender/Sex-Based Bias Score: 0. There is no content in this text that implies one sex or gender is inferior to another, incites or encourages abuse or violence against someone because of their sex, gender, or sexual orientation, or is demeaning, objectifying, or threatening based on gender or sex.

Religious Bias Score: 0. There is no content in this text that promotes beliefs of religious superiority or inferiority, upholds stereotypes or biases about specific religions, or associates religions with negative behaviors or characteristics.

Ability-Based Bias Score: 0. There is no content in this text that disparages someone's abilities or disabilities, attributes abilities or disabilities with intent to insult, or upholds stereotypes and biases related to abilities or disabilities.

Origin-Based Bias Score: 0. This text uses the term ``"Chinaman"", which is now considered outdated and offensive, but the text does not use it to express harmful ideas or opinions about people from China.

Violence Score: 0. There is no aggressive or violent content in this text.

\section{Appendix C. LLM Annotation Prompt} \label{app:prompt}
\setcounter{section}{3}

\textbf{System Prompt}
You are an expert artificial intelligence annotator for toxicity and harmfulness. You are provided the following categories of toxicity:

Racial Discrimination: promotes or suggests racial superiority/inferiority; perpetuates harmful racial stereotypes; or promotes racially motivated insults, attacks, or crimes. Racial discrimination is distinct from discrimination based on someone’s country of origin, religion, or language. 

Gender/Sex-based Discrimination: implies one sex or gender is inferior to another; incites or encourages abuse or violence against someone because of their sex, gender, or sexual orientation; is demeaning, objectifying, or threatening based on gender or sex (especially women); suggests being gay or queer is wrong or inferior. 

Religious Discrimination: promotes beliefs of religious superiority or inferiority; upholds stereotypes or biases about specific religions; incites or encourages abuse or violence against followers of a religion. 

Ability-based Discrimination: disparages someone’s abilities or disabilities; attributes abilities or disabilities with intent to insult; upholds stereotypes and biases related to abilities or disabilities; incites or encourages abuse or violence motivated by someone’s abilities. 

Origin-based Discrimination: upholds or promotes stereotypes related to nationality, place of origin, or immigration status (especially immigrant or refugee status); incites or encourages abuse or violence motivated by someone’s origin. 

Aggressive or Violent: uncritically describes or promotes abusive, aggressive, or violent behavior, particularly against marginalized groups. 

Annotation problem: Given a piece of text, provide a score between 0 and 3 for each of the proposed categories based on the severity of each. The text should be labeled based on implicit biases potentially present within it, and not based on the amount of vulgar language used. 

Rate each text with a score from 0-3, using this scale exactly:

0 - text mentions stereotypes about a particular group of people or acts of violence, but does so in an objective way; text does not promote biases against any group or acts of violence.

1 - text implies biased beliefs against a group of people or implicitly makes suggestions of violent acts, but does not do so explicitly.

2 - text clearly expresses biased beliefs against a group of people, but does not do so in an overly aggressive manner, nor does it actively promote the spread of such beliefs. 

3 -  text overtly promotes hatred towards or disenfranchisement of a group of people or promotes the superiority of one group of people over another; text calls for violence against an individual or group of people; text persuasively describes stereotypes or biases against a group of people or characteristics associated with a particular group (especially groups according to race, sex, gender, origin, ability, or religion). 

Always respond in this format exactly:

\#\# Racial Discrimination Score \#\# :  A score from 0 to 3

\#\# Racial Discrimination Reasoning \#\# : What does this text contain that makes it racially discriminatory? Provide specific examples and detailed reasoning.

\#\# Gender/Sex-Based Discrimination Score \#\# :  A score from 0 to 3

\#\# Gender/Sex-Based Discrimination Reasoning \#\# : What does this text contain that makes it discriminatory based on gender or sex? Provide specific examples and detailed reasoning

\#\# Religious Discrimination Score \#\# :  A score from 0 to 3

\#\# Religious Discrimination Reasoning \#\# : What does this text contain that makes it religiously discriminatory? Provide specific examples and detailed reasoning.

\#\# Ability-Based Discrimination Score \#\#:  A score from 0 to 3

\#\# Ability-Based Discrimination Reasoning \#\#: What does this text contain that makes it discriminatory based on ability? Provide specific examples and detailed reasoning.

\#\# Origin-Based Discrimination Score \#\#:  A score from 0 to 3

\#\# Origin-Based Discrimination Reasoning \#\#: What does this text contain that makes it discriminatory based on origin? Provide specific examples and detailed reasoning.

\#\# Aggressive or Violent Score \#\#:  A score from 0 to 3

\#\# Aggressive or Violent Reasoning \#\#: What does this text contain that creates aggressive or violent themes? Provide specific examples and detailed reasoning. 

\newpage
\section{Appendix D. Annotation Scores Distribution by Dataset} \label{app:score_distribution}
\setcounter{section}{4}
\setcounter{figure}{0}

Proportion of annotation scores from Llama 3.1 across all datasets, not including scores of 0. This reflects the distribution of scores across the texts, which were sampled to create the classifier training data.

\begin{figure}[h!]
    \centering
    \includegraphics[width=0.6\linewidth]{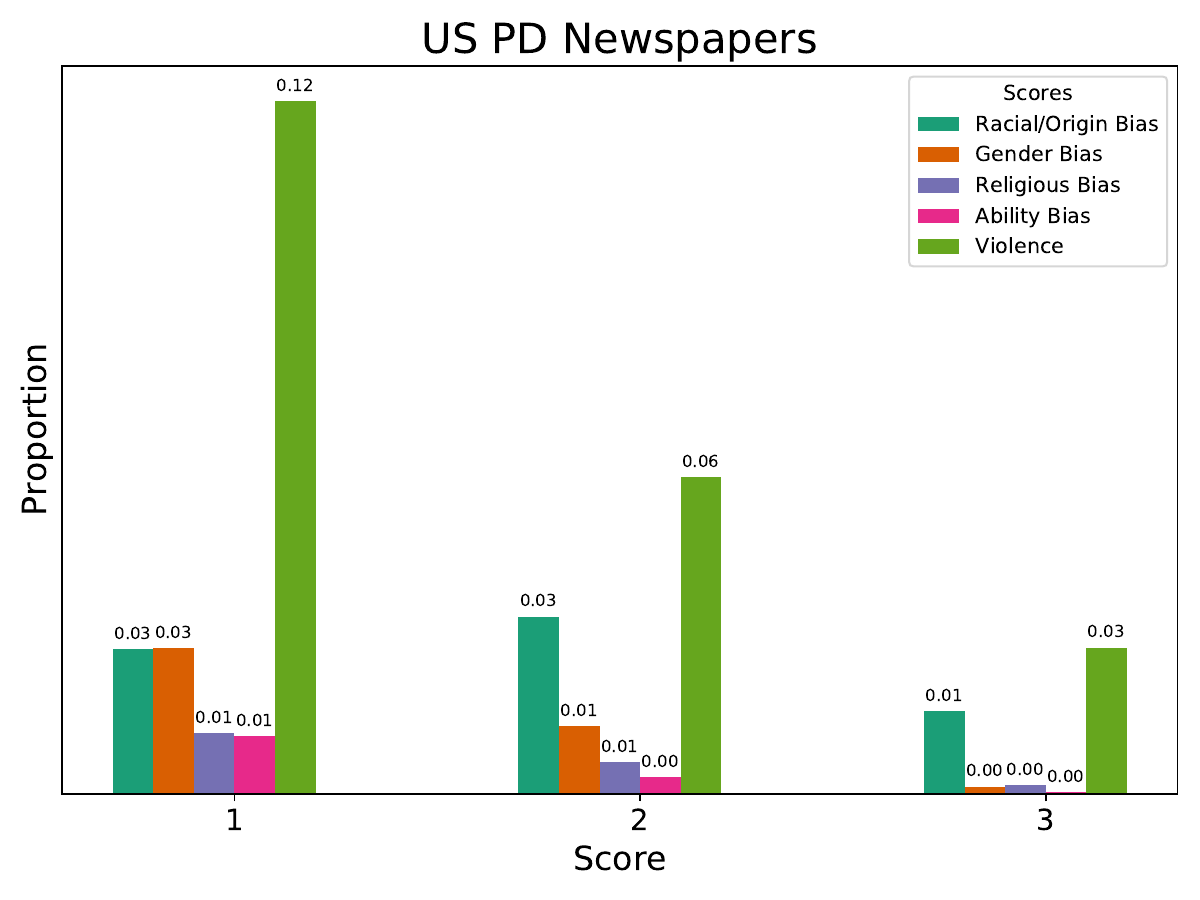}
\end{figure}

\begin{figure}[h!]
    \centering
    \includegraphics[width=0.6\linewidth]{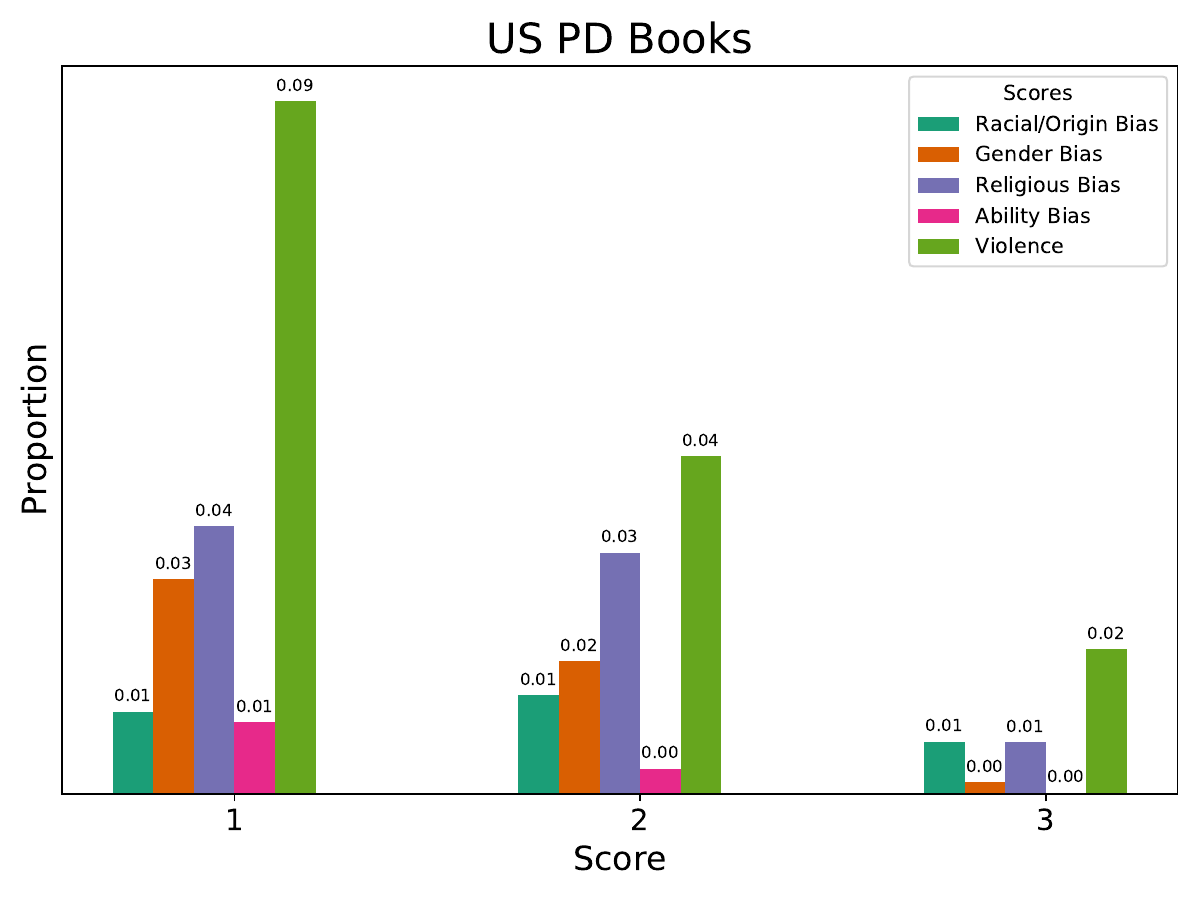}
\end{figure}

\begin{figure}[h!]
    \centering
    \includegraphics[width=0.6\linewidth]{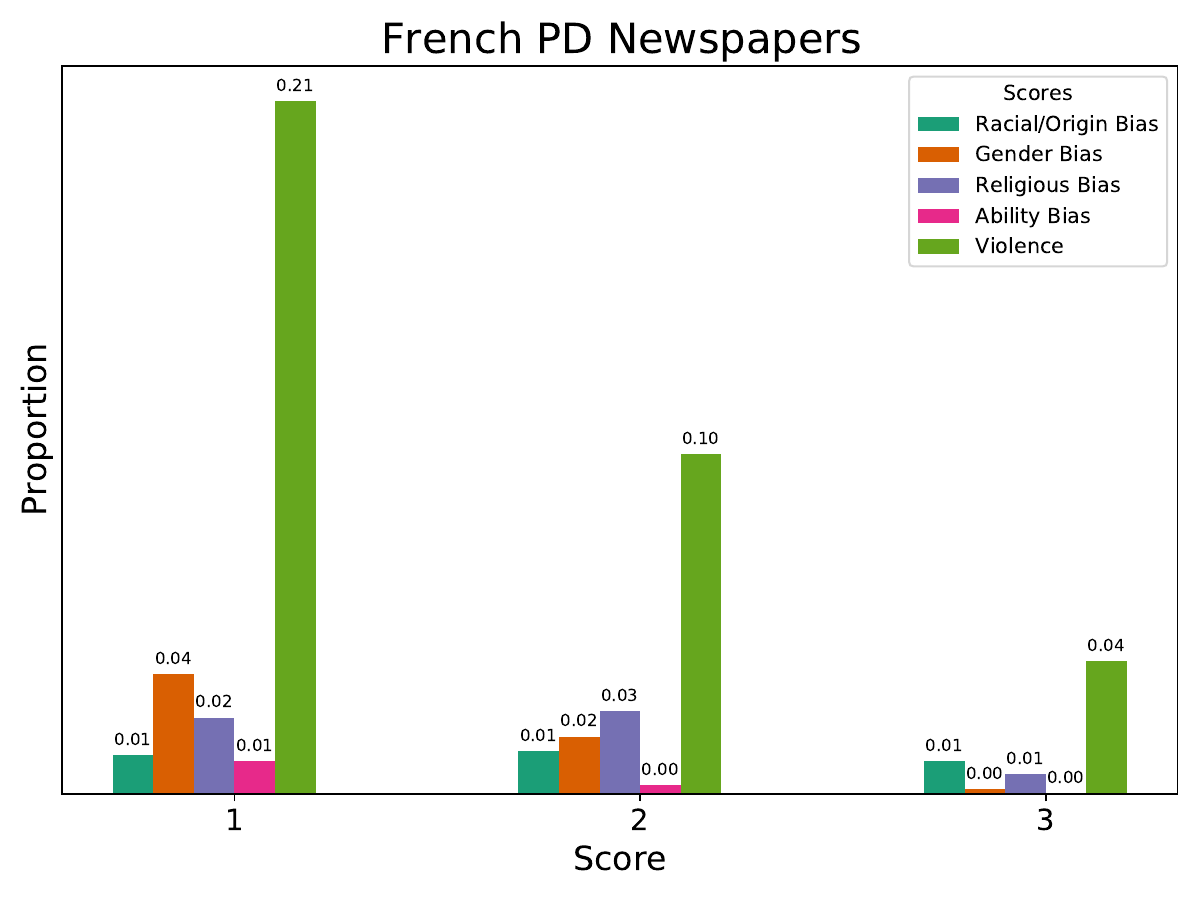}
\end{figure}
\newpage
\begin{figure}[h!]
    \centering
    \includegraphics[width=0.6\linewidth]{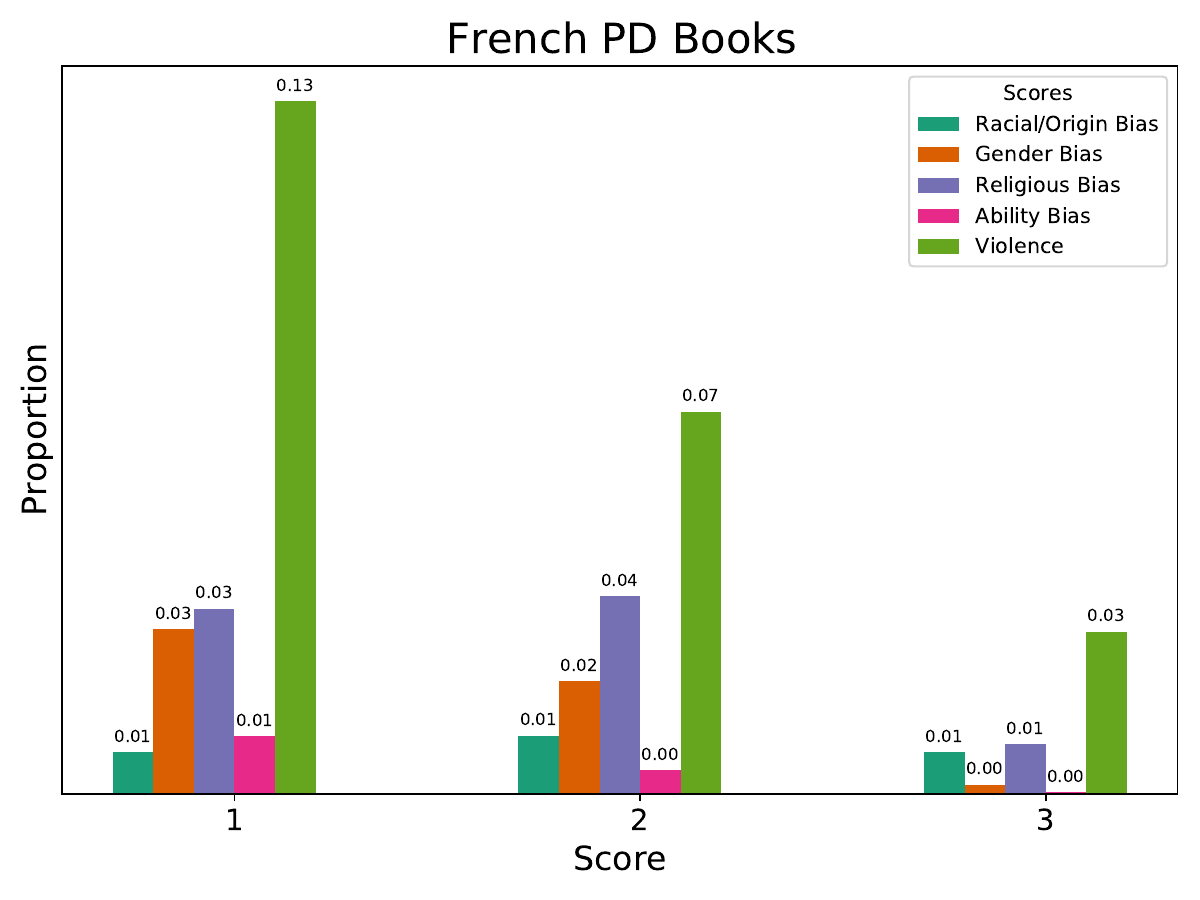}
\end{figure}

\begin{figure}[h!]
    \centering
        \includegraphics[width=0.6\linewidth]{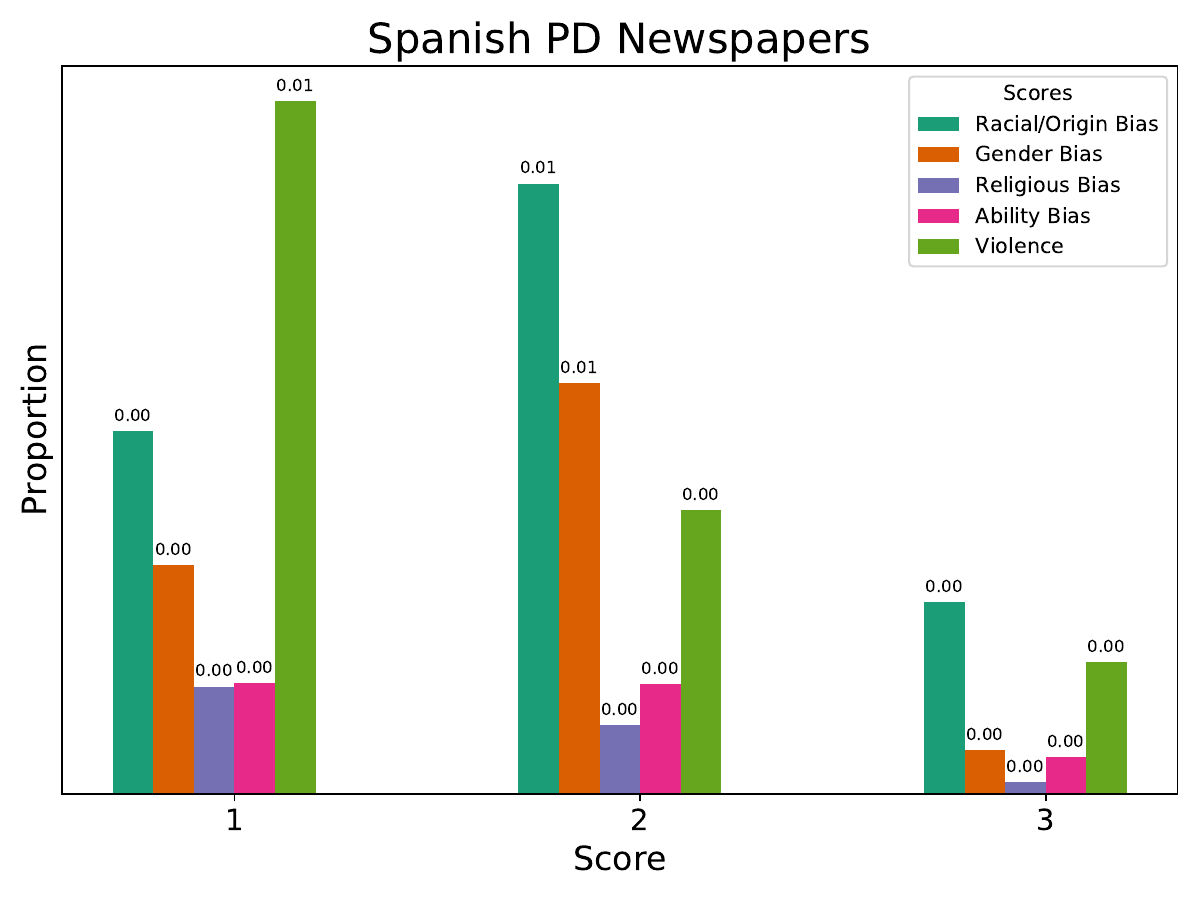}
\end{figure}
\newpage
\begin{figure}[h!]
    \centering
        \includegraphics[width=0.6\linewidth]{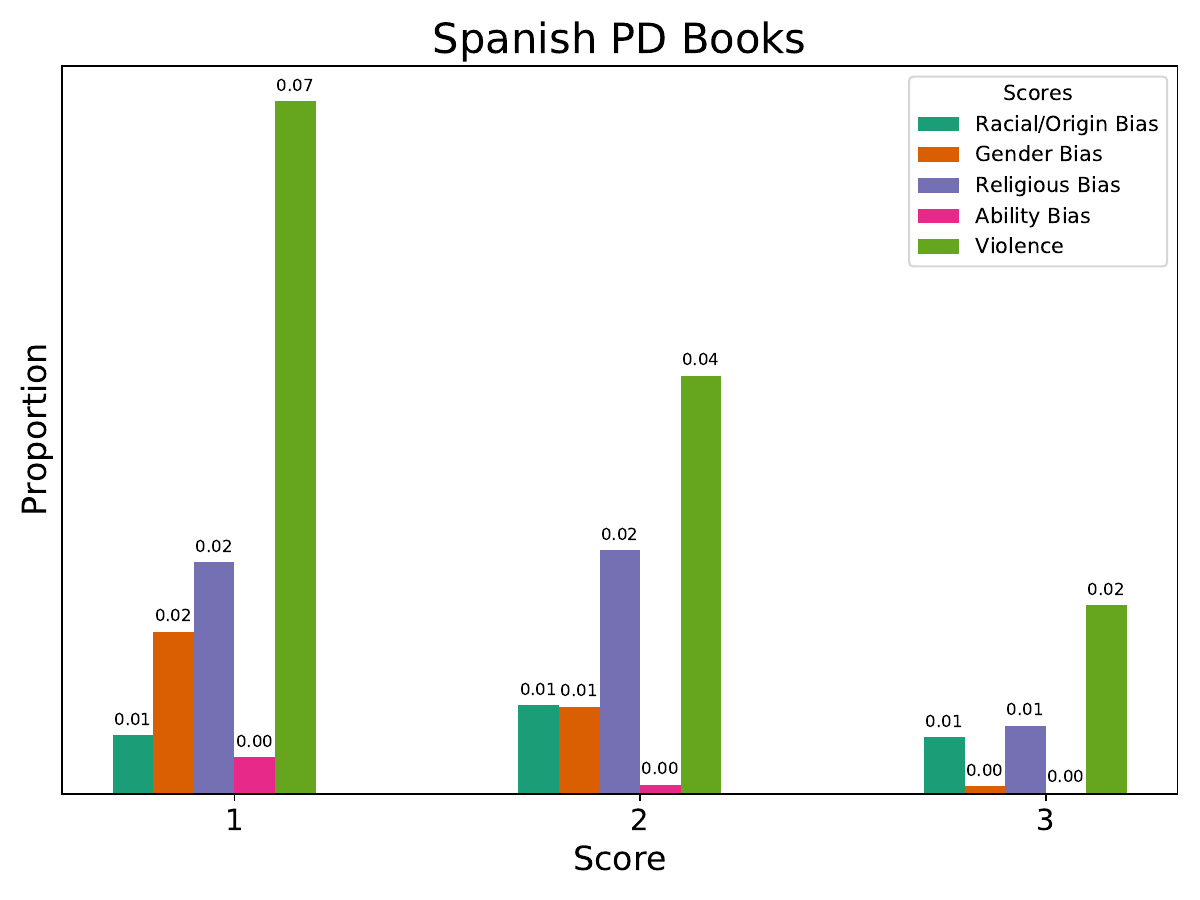}
    \caption{}
    \label{fig:german_scores}
\end{figure}

\begin{figure}[h!]
    \centering
    \includegraphics[width=0.6\linewidth]{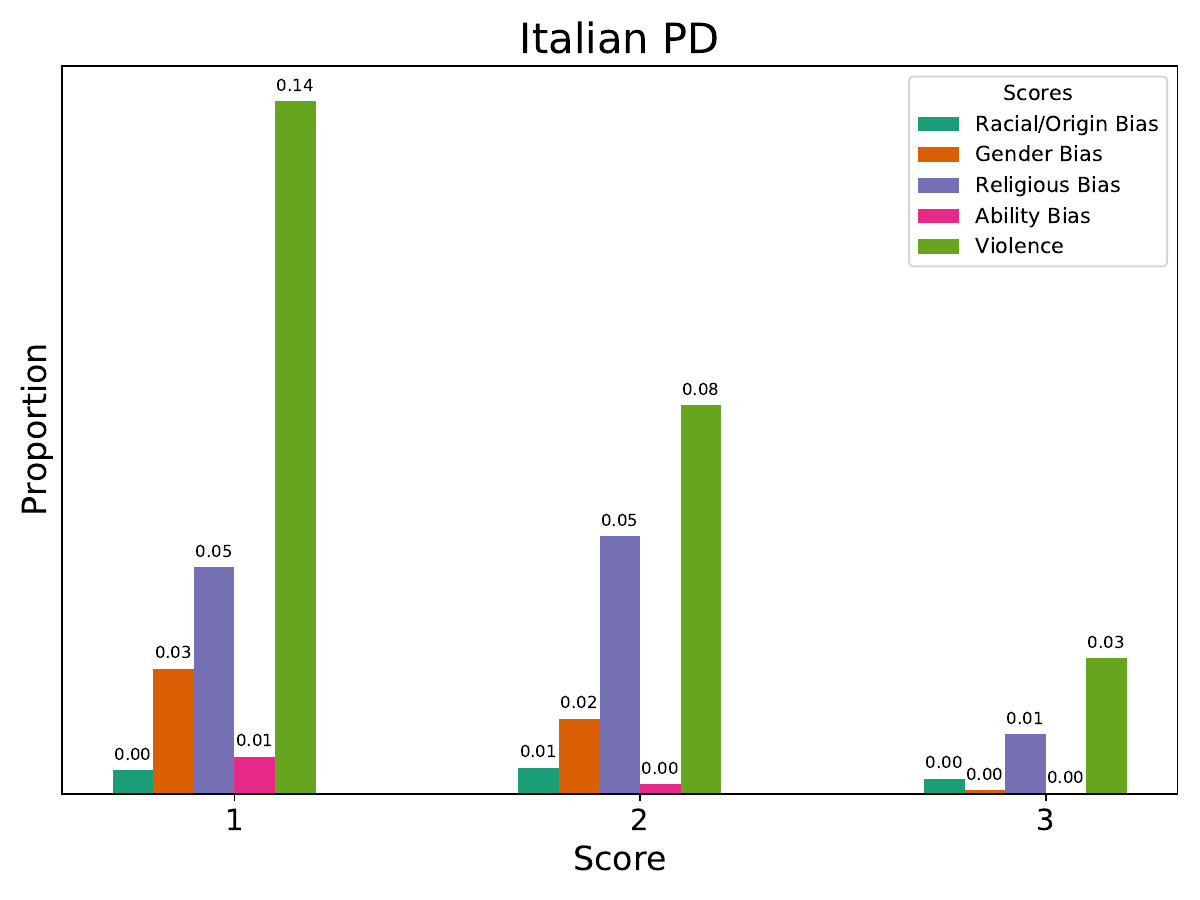}
\end{figure}
\newpage
\begin{figure}[h!]
    \centering
    \includegraphics[width=0.6\linewidth]{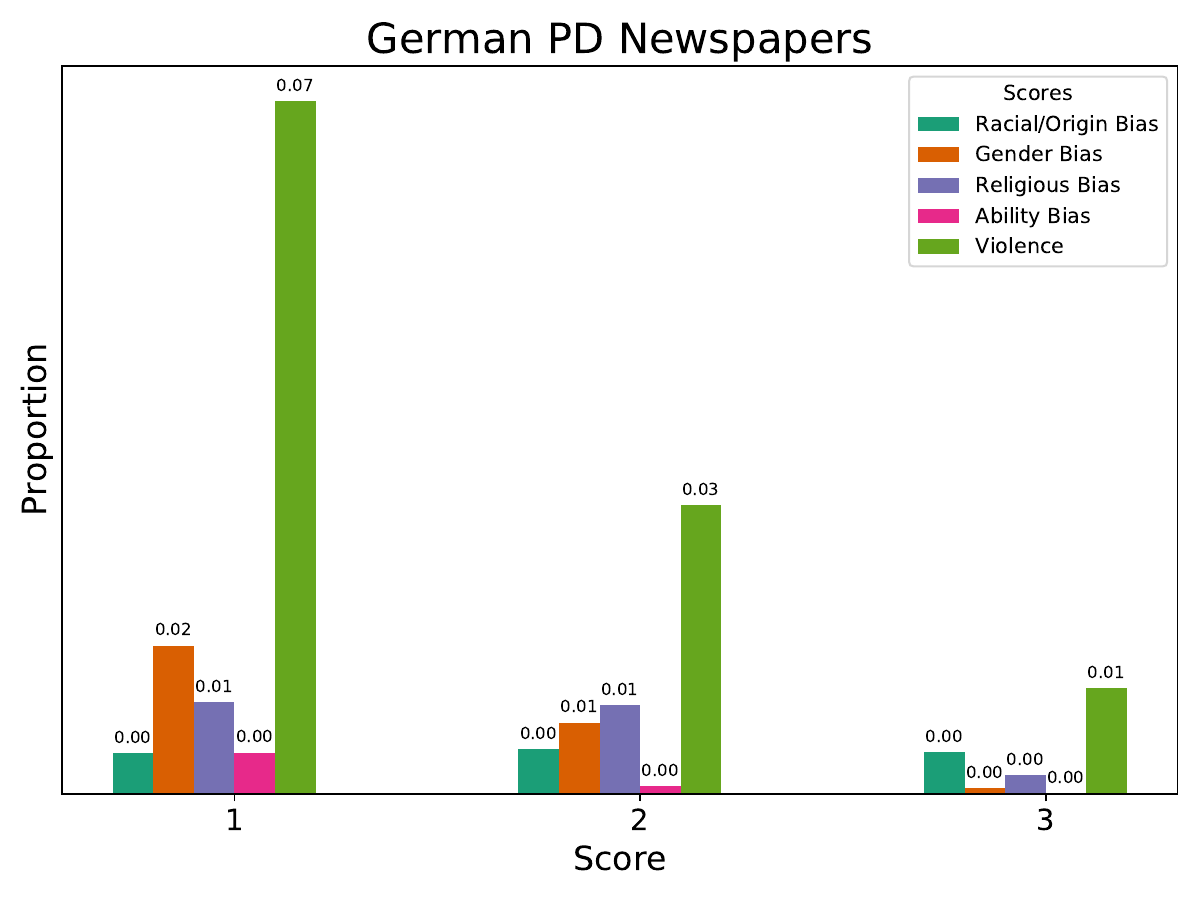}
\end{figure}

\begin{figure}[h!]
    \centering
    \includegraphics[width=0.6\linewidth]{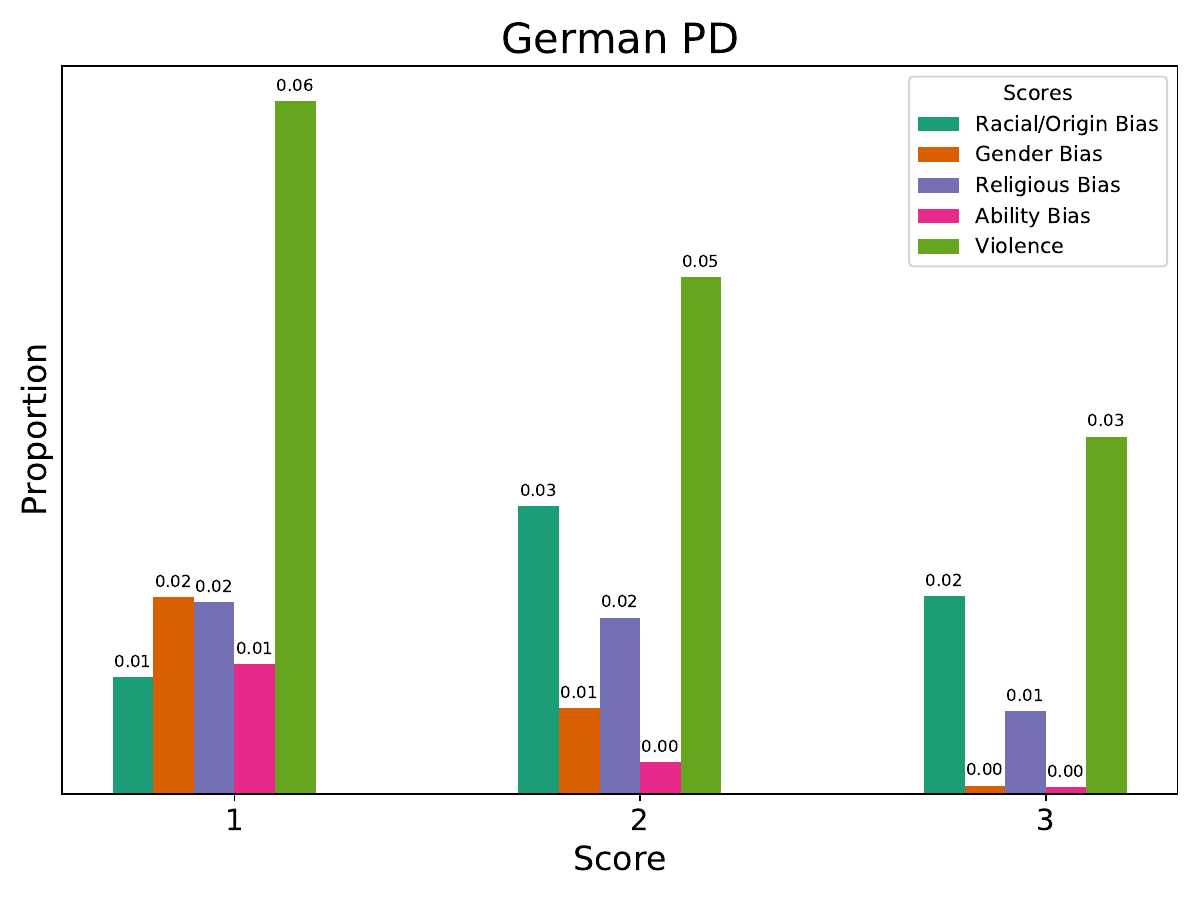}
\end{figure}
\newpage
\begin{figure}[h!]
    \centering
    \includegraphics[width=0.6\linewidth]{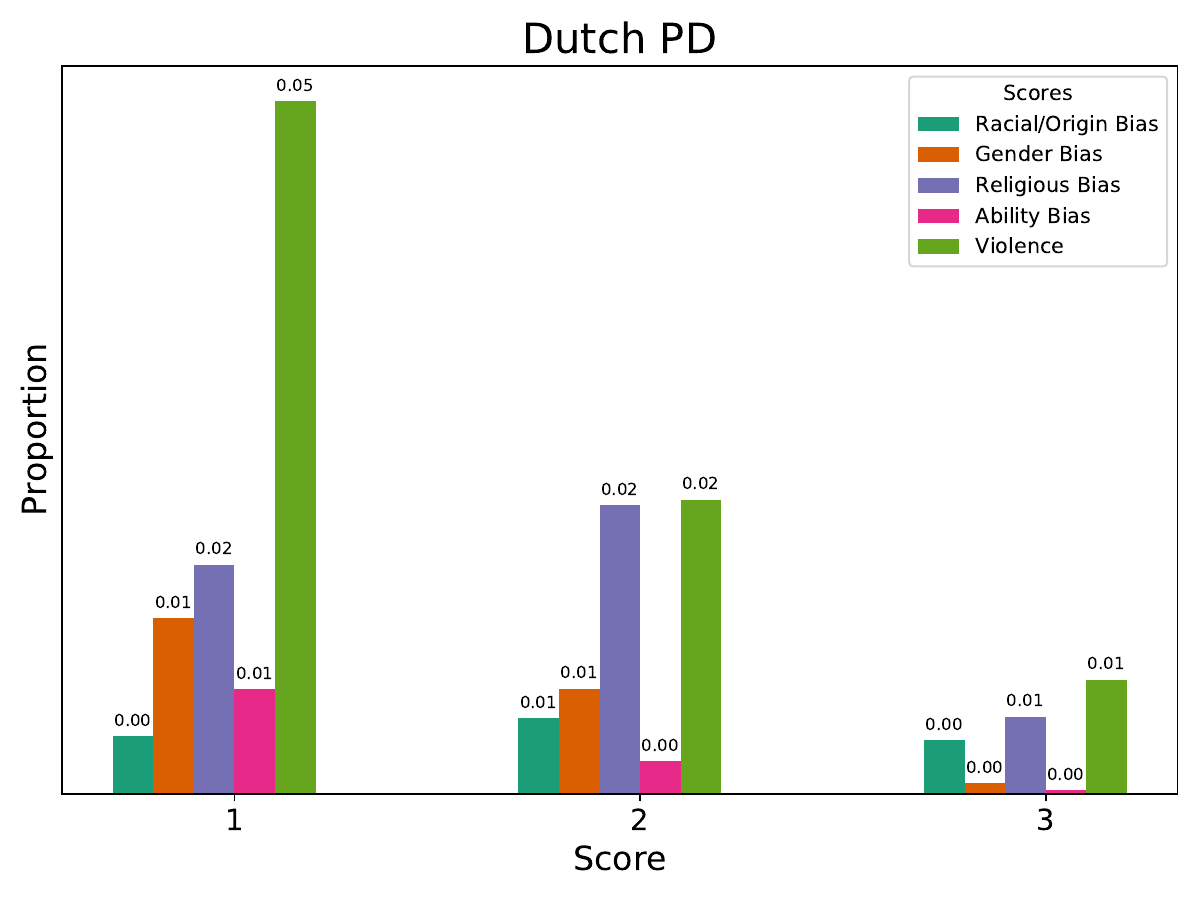}
\end{figure}

\begin{figure}[h!]
    \centering
    \includegraphics[width=0.6\linewidth]{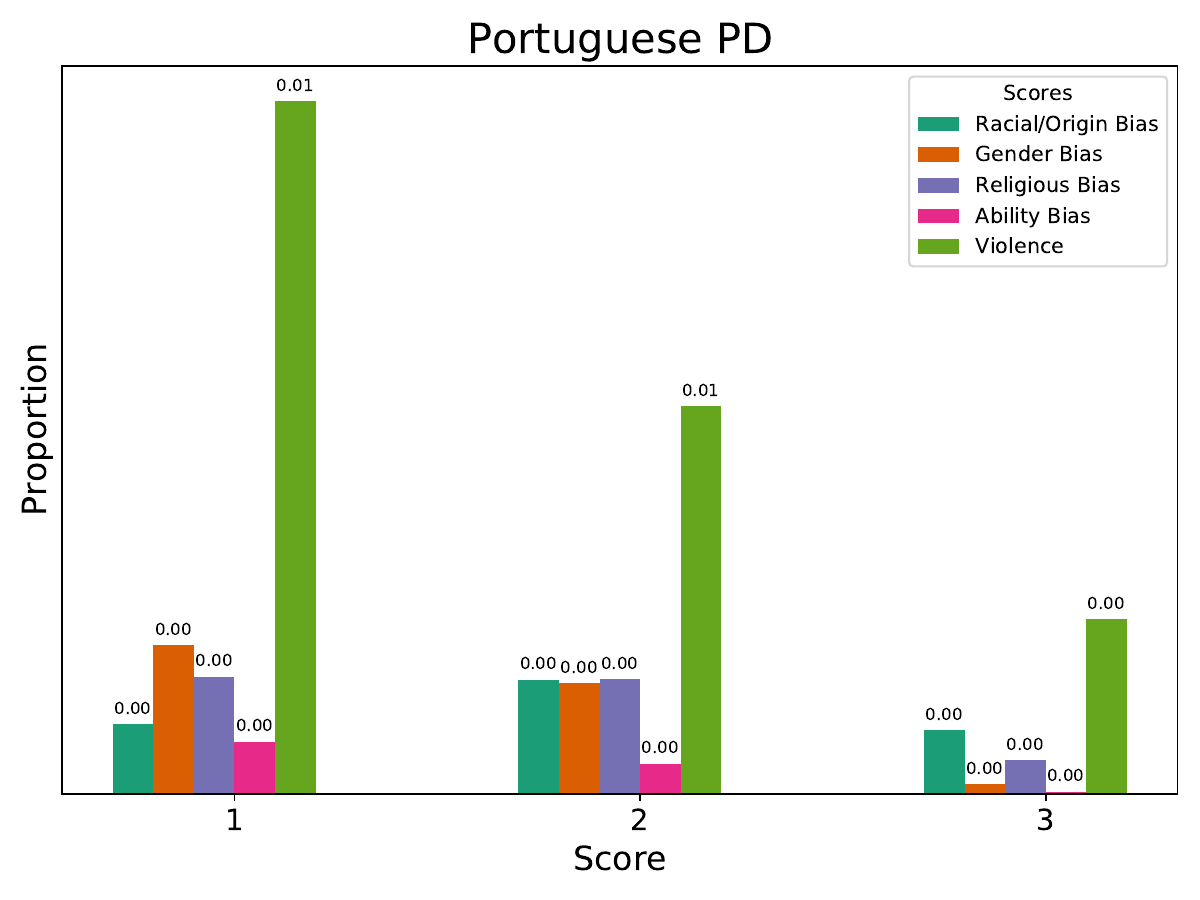}
\end{figure}

\newpage
\begin{figure}[h!]
    \centering
    \includegraphics[width=0.6\linewidth]{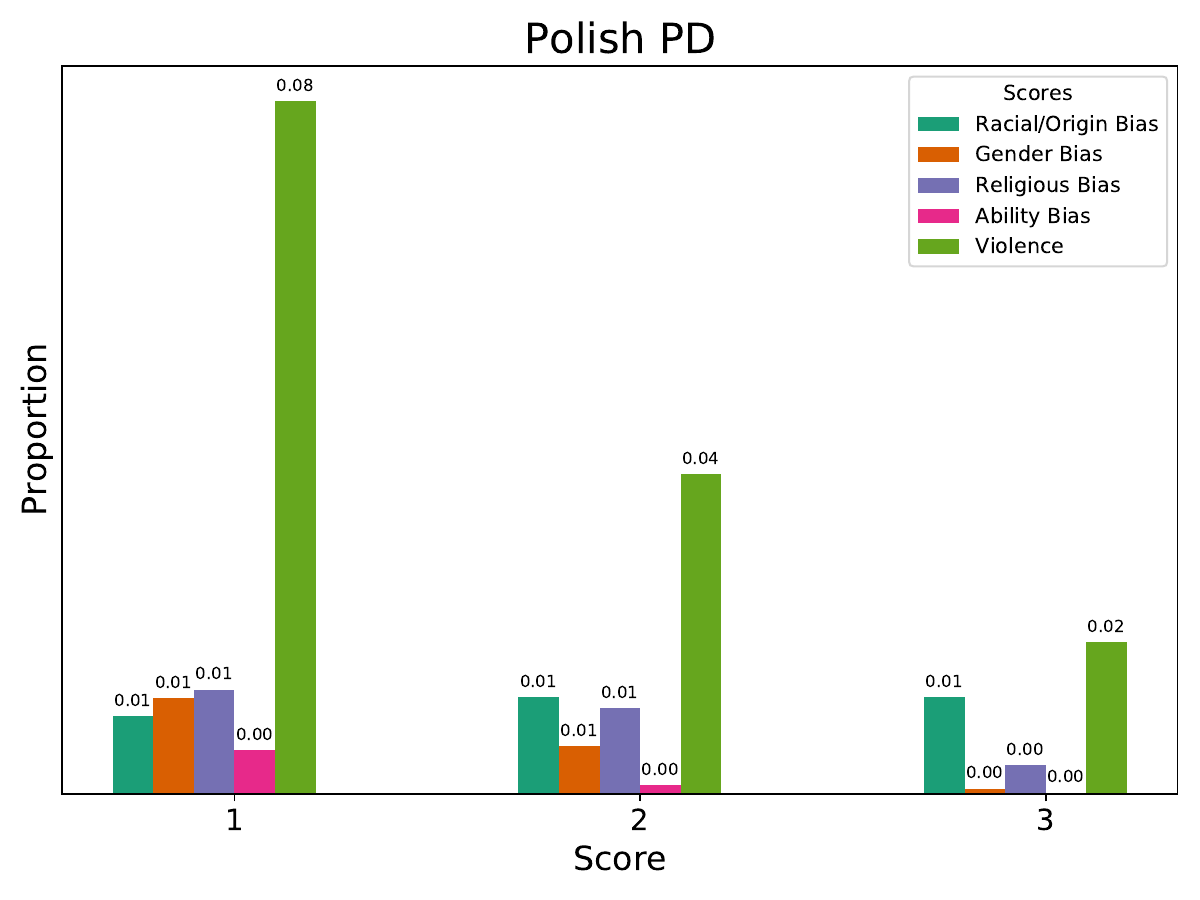}
\end{figure}

\begin{figure}[h!]
    \centering
    \includegraphics[width=0.6\linewidth]{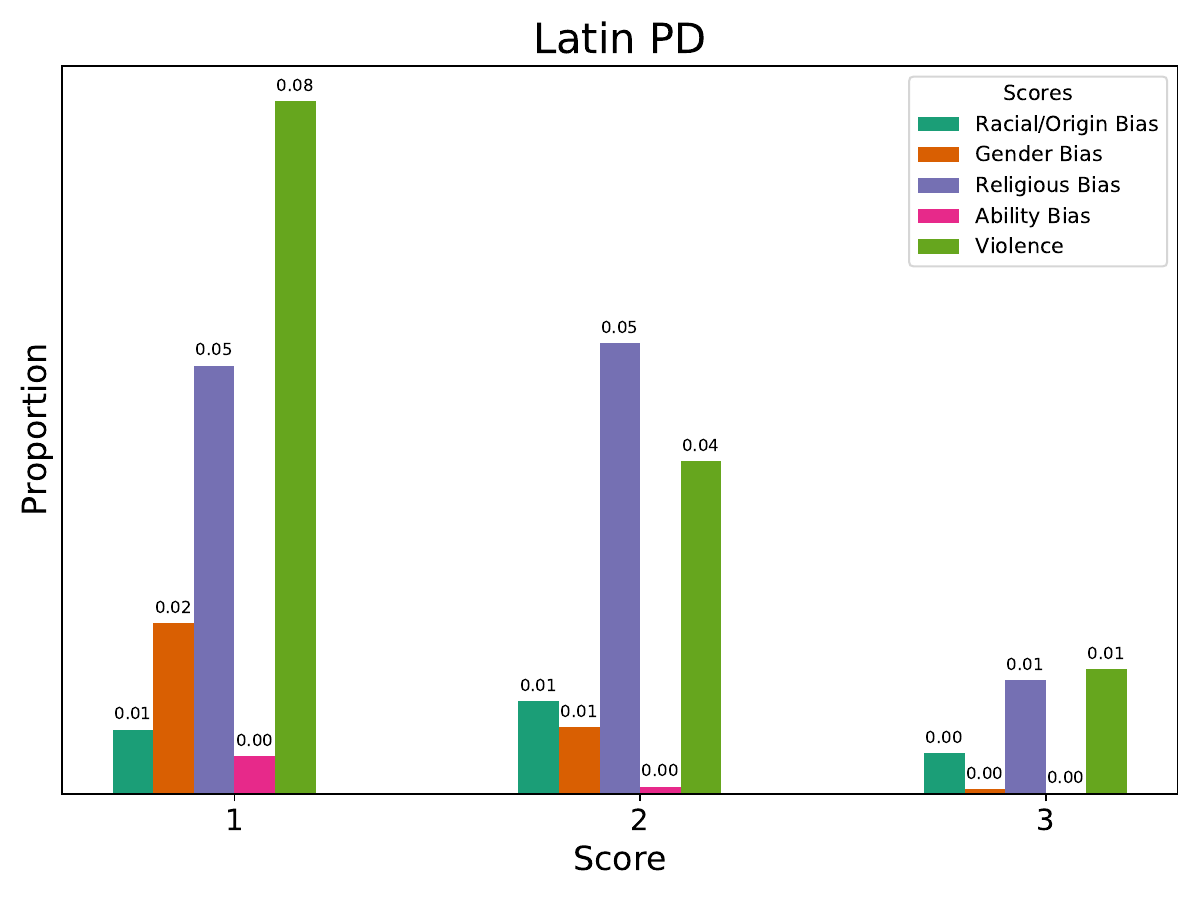}
\end{figure}

\newpage
\section{Appendix E. Comparing Annotations Across models}
\label{app:model_comparison}
\setcounter{section}{5}
\setcounter{table}{0}

In order to determine the model whose internal behaviors aligned most with what we desired as outputs from the annotations, we compared three different models: GPT-3.5, Claude 3 Haiku, and Llama 3.1. The below samples compare the results of the models to our human annotations. 


\smallskip

\textbf{Text:} “Tvkoa. Wash., March 23. Four drunken [MASK] from the Coeur d'AK-ne reservation went Into Whltmore and McLean's grocery store lat night. After using noisy and abusive language they began to throw crockery and mer chandise around the room. Richard Kelly came to the aid of McLean and was knocked down three times by the Infuriated [MASK]. He seized a large hammer and knocked out two of the red men. Constable Oliver Sparks, hearing the noise, came on from the street, and with the weights from the scales and counter knocked out the other two. All four received ugly cuta on the head, and iit was feared for some time that thy had been killed. Physicians were summoned and their wounds d rested so that they could be taken to the calaboose. The citizens are aroused and are determined to send the offenders to the penitentiary. PASSED THE ASSEMRLY. Albany, N. Y., March 23. The Great er New York charter bill passed the assembly by a vote of 118 to 2S. It eorqes uo In the senate tomorrow. PAWNBROKER ARRESTED. Yesterday a [MASK] stole from the rooms of a white woman a lot of silk garments and other artlclea. The woman was formerly married to a [MASK], and had a number of fine Chinese suits. Officer Oberg waa notified, and at once started on the track of the thief. Happening into the pawnshop of Ark Wo on Bond street." 

\begin{table}[h!]
\centering
\begin{tabular}{|l|c|c|c|c|c|c|}
\hline
Category & R & G & Re & A & O & V \\
\hline
Human         &    0 &      0 &        0 &       0 &      2 &        0 \\ 
\hline
LLaMa 0-SHOT  &    3 &      0 &        0 &       0 &      3 &        2 \\ 
\hline
GPT3.5 0-SHOT &    3 &      0 &        0 &       0 &      0 &        2 \\ 
\hline
Claude 0-SHOT &  N/A &    N/A &      N/A &     N/A &    N/A &      N/A \\ 
\hline
LLaMa 1-SHOT  &    3 &      0 &        0 &       0 &      2 &        2 \\ 
\hline
GPT3.5 1-SHOT &    2 &      0 &        0 &       0 &      0 &        1 \\ 
\hline
Claude 1-SHOT &    3 &      0 &        0 &       0 &      0 &        2 \\ 
\hline
LLaMa 2-SHOT  &    3 &      0 &        0 &       0 &      3 &        3 \\ 
\hline
GPT3.5 2-SHOT &    2 &      0 &        0 &       0 &      0 &        2 \\ 
\hline
Claude 2-SHOT &    3 &      0 &        0 &       0 &      2 &        2 \\ 
\hline
LLaMa 3-SHOT  &    3 &      0 &        0 &       0 &      3 &        2 \\ 
\hline
GPT3.5 3-SHOT &    2 &      0 &        0 &       0 &      0 &        2 \\ 
\hline
Claude 3-SHOT &    2 &      0 &        0 &       0 &      2 &        3 \\ 
\hline
\end{tabular}
\caption{Comparison of Human Ground Truth with 0-SHOT, 1-SHOT, 2-SHOT, and 3-SHOT Results Across Models. R: Race, G: Gender, Re: Religion, A: Ability, O: Origin, V: Violence.}
\label{tab:comp1}
\end{table}

\newpage
\textbf{Text:} “In any of tho cotton, rice, or sugar growing districts of tho South, tho very same relations between tho whites and blacks may bo found that this correspondent has reported from South Carolina. In tho cities, all tho servants, porters, cartmou, stevedores, etc., are blacks; in tho country, all tho flold-honds are blacks. Tho whites may koop tho shops, onlar tho professions, hold tho offices, and hang around tho central portions of tho town talking politics, but they are never found in any vocation which roqaires manual labor. There aro those who have been (aught that such work carries disgrace with it, and is fit only for [MASK] tboro oro others who profess this doctrine merely ns an excuse for their refusal to work, and still others who havo no uxoaso but ‚Äö$\sqrt{}$Ñ$\sqrt{}$$\int$pure cassodnuss."" All agree, however, in tho nbsolatoandnnivorsal determination not to soil their hands by manual labor, no matter if poverty stares them in tho faco. No community can afford to support more than a certain proportion of drones, and the proportion in all Southern communities is too largo. In any county or town where there are 1,000 negroes, there may almost always bo found the same number of idle whites. The decline of prosperity in such a community is a more question of time. 1 The blacks cannot if they would, aud will not if they could, carry the whole burden of supporting themselves aud an equal number of lazy whites who affect o superiority over them.”

\begin{table}[h!]
\centering
\begin{tabular}{|l|c|c|c|c|c|c|}
\hline
Category & R & G & Re & A & O & V \\
\hline
Human         &     2 &       0 &         0 &        0 &       0 &         0 \\
\hline
LLaMa 0-SHOT  &     2 &       0 &         0 &        0 &       0 &         0 \\
\hline
GPT3.5 0-SHOT &     1 &       0 &         0 &        0 &       0 &         0 \\
\hline
Claude 0-SHOT &     3 &       0 &         0 &        0 &       0 &         1 \\
\hline
LLaMa 1-SHOT  &     3 &       0 &         0 &        0 &       0 &         0 \\
\hline
GPT3.5 1-SHOT &     2 &       0 &         0 &        0 &       0 &         0 \\
\hline
Claude 1-SHOT &     3 &       0 &         0 &        0 &       0 &         0 \\
\hline
LLaMa 2-SHOT  &     3 &       0 &         0 &        0 &       3 &         0 \\
\hline
GPT3.5 2-SHOT &     3 &       0 &         0 &        0 &       0 &         0 \\
\hline
Claude 2-SHOT &     3 &       0 &         0 &        0 &       0 &         0 \\
\hline
LLaMa 3-SHOT  &     3 &       0 &         0 &        0 &       2 &         0 \\
\hline
GPT3.5 3-SHOT &     3 &       0 &         0 &        0 &       0 &         0 \\
\hline
Claude 3-SHOT &     3 &       0 &         0 &        0 &       0 &         0 \\
\hline
\end{tabular}
\caption{Comparison of Human Ground Truth with 0-SHOT, 1-SHOT, 2-SHOT, and 3-SHOT Results Across Models. R: Race, G: Gender, Re: Religion, A: Ability, O: Origin, V: Violence.}
\label{tab:comp2}
\end{table}

Note that a score of "N/A" means that the model refused to provide a response for the particular text. Further, for the final results we chose to merge race-based discrimination and origin-based discrimination into one category.

\newpage
\section{Appendix F. Example LLM Annotations} \label{app:example_annotations}
setcounter{section}{6}

All examples come from US PD Newspapers. Each of these texts has a rating of 3 for a given dimension. 

\subsection{High Racial Bias Score}

\textbf{Text:}

\begin{quote}
    This table is made up to cover only the higher officers in all departments. The legions of subalterns are nearly all from north of the line. Indeed, whenever a southern member of Congress gets a place for a southern man or woman, it is such a marked event that it gets into the newspapers. This record gives to the north 1,977 heads of departments and commissioned officers, and only 173 to the south. It gives to the north \$5,500,000 and to the south less than half a million in salaries per annum. This is but a sample of the way in which our paternal government treats the south. This accounts for only six millions out of nearly a billion that is spent. Of the rest, we get nothing to speak of, outside of the pay to postmasters, members of Congress, and revenue collectors. One hundred and fifty million dollars goes to northern pensions and several millions to northern cemeteries. Of course, Mr. Harrison will say that he would have given more places to the south, but he couldn't find white men in the southern republican party that were fitted to hold them. That is the reason, I reckon, why he gave some of the best offices to the negroes; but there was no law against his appointing conservative democrats. 
\end{quote}

\textbf{LLM Annotation:}

Racial/Origin-Based Discrimination Score: 3

Racial/Origin-Based Discrimination Reasoning: The text contains overtly discriminatory language and sentiments against people from the southern region of the United States. It implies that individuals from the north are inherently more qualified and deserving of positions of power than those from the south. The statement ``Mr. Harrison will say that he would have given more places to the south, but he couldn't find white men in the southern republican party that were fitted to hold them" suggests that the author believes southerners are less capable or inferior due to their geographical origin. Furthermore, the implication that the best offices were given to ``negroes" instead of conservative democrats is a clear example of racial bias, as it perpetuates stereotypes and prejudices against African Americans.

\bigskip
\textbf{Text:}

\begin{quote}
    They loved it well enough to fight for it. They never gave it up until every battlefield upon which they could muster an army was red with their blood. But they lost their homes, nevertheless — why? Because in the subtler combat of mind against mind they were no match for the whites. The pale face deceived his red brother, when the Indians were the strongest; and when at length the whites were the stronger, the red men had to give up their homes.

    Farmers Deprived of Birthright.
    Brethren of the South, will you learn nothing from the past? Have you no eyes to see what is going on? Do you not realize that in the war of wits you are losing ground? Will you never understand that national politics and laws can be so shaped as to give all the advantage to one class, or one section? Is it impossible for you to learn that special privilege always lives at the expense of the unprivileged—is a deadly parasite that will sap the life of the noblest tree?
    
    Use your eyes. Look about you. See things as they are. Where is the bulk of the wealth of the nation? In that portion of it which nature did the least favor. How did bleak, barren New England come to be so rich? She made the laws to suit herself, and these laws took the prosperity of the South and West and gave it to the capitalists of the East and North.

Who owns your railroads? The North. Your mills? The North.
\end{quote}

\textbf{LLM Annotation:}

Racial/Origin-Based Discrimination Score: 3

Racial/Origin-Based Discrimination Reasoning: This text contains overtly discriminatory language and sentiments against Native Americans, portraying them as inferior to whites in the ``combat of mind against mind." It suggests that Native Americans were deceived by whites when they were stronger, implying a lack of intelligence or foresight on the part of the Native Americans. The text also uses stereotypes such as ``red men" and implies that Native Americans are incapable of competing with whites in terms of wealth and prosperity. Furthermore, it presents a simplistic and inaccurate view of historical events, blaming the loss of their homes on the supposed intellectual superiority of whites.

\subsection{High Gender Bias Score}

\textbf{Text:}

\begin{quote}
    Every female inmate, who changes her mind, and wants to get out, and marry, and be the mother of a family, should be given that privilege. No true religion demands the incarceration of the unwilling captive. No church, truly Christian, would stifle the legitimate promptings of God Almighty. To say that piety demands the darkness of a dungeon, is the same as saying that the rose blooms best in the cellar. To say that the most unnatural life is the most godly, is the same as saying God didn’t know what He was doing when He made the laws of nature. Let the State enter all these places where her citizens are held, and put these questions: (1.) How came you here? (2.) Did you do it? (3.) Do you want to get out? The questions should not be heard by the jailers, or by any one representing the guilty Roman church. Their presence would amount to intimidation, and would defeat the object the State has in view. To those inmates who say they want to get out, immediate freedom should be given. Otherwise, they will suddenly die, or disappear. The officers of the State should be empowered to take those inmates away with themselves, when they leave the premises. Never trust a priest, in such a matter, one single minute.
\end{quote}

\textbf{LLM Annotation:}

Gender/Sex-Based Discrimination Score: 3

Gender/Sex-Based Discrimination Reasoning: This text contains discriminatory content based on gender, as it suggests that women are primarily valued for their roles within families (e.g., being ``the mother of a family"). It implies that women's primary value lies in their domestic and reproductive capabilities, reinforcing stereotypes about women's roles and abilities. Furthermore, the text assumes that women who want to leave prison and get married are doing so because they want to fulfill traditional feminine roles, rather than considering alternative reasons for their desire to leave prison. This reinforces patriarchal attitudes towards women and their agency.

\bigskip
\textbf{Text:}
\begin{quote}
    Take the perfectly formed woman, and her head is distinctly feminine, her hair long and silky; her neck is long, curved, graceful, swan-like; her face is made for beauty, as man's is for strength; her lips are a cupid’s bow, inviting kisses, while her teeth are small, pearly; her bosom suggests motherhood and the sustenance of her babe—a blow there would kill her; the shoulders are narrow, the arms and hands soft, delicate; her hips are broad, suggesting the broadening necessary to child-bearing; her thighs are most tender, and her leg tapers to a small foot, never meant for the rougher, rocky path in life; her voice reveals her sex as far as it can be heard; her spirit is that of a dependent queen, yearning for a king; and she is happiest when fitly mated, ruling her household, rearing her children, honoring the monarch of her home.
    Now, if the lioness had been designed for the same purpose as the lion, Nature would not have created them upon such different plans. Whenever the Creator made a difference, something was meant by it. Accidents cannot happen to the Creator. Mistakes cannot be made by the All-wise, and All-seeing, and All-powerful.
    A man may be quite sure that if Noah carried several pairs of all quadrupeds and other animals into that little boat of his, the female cows, goats, horses, gorillas, buffaloes, and elephants differed as much from the males, as they do at this time.
    The most ancient pictures, statues, tombs, skeletons, and mummies that the indefatigable diggers and explorers have found, show us precisely the same difference in sexes that we now see.
No tribe of human beings has ever been discovered where the spheres of action for men and women were the same.
\end{quote}

\textbf{LLM Annotation:}

Gender/Sex-Based Discrimination Score: 3

Gender/Sex-Based Discrimination Reasoning: This text explicitly promotes a stereotypical view of women as ``dependent queens," who yearn for a king and are happiest when mated and ruling their household. It suggests that women are inherently designed for domestic duties and are less capable of navigating the ``rougher, rocky path in life." The text implies that men and women have different spheres of action and that accidents or mistakes cannot be made by the Creator, reinforcing the idea that this is an inherent and natural distinction between the sexes.

\subsection{High Religious Discrimination Score}

\textbf{Text:}

\begin{quote}
    The Alaskan Missionary
    
    An Unnecessary Evil
    
    A. Franklin Lesslie
    
    O King, from the time of the Egyptian Pharos, down through the centuries to the present day, ever ruled with a more despotic sway than the Alaskan missionary of today rules over the simple-minded Siwash Indians, living in the interior of Alaska, and it is with regret I make the assertion, that the ruling is to the detriment, and not the betterment, of the Indian.
    I want nothing in common with a religion, or any other thing, which plays upon the superstitions of a simple-minded people to gain their allegiance, and I have less in common with the Catholic church than I have with any other religion or belief that exists, either in Alaska or out of it.
    The Roman Catholic religion, as represented by the Jesuit priests in the interior of Alaska, has for its representation the most profligate collection of parasites (a few excepted) with which it is possible to burden a community, either religious or civil, and little wonder it is that the pious old libertines, who button their haberdashery in the back, receive scant consideration at the hands of the big-hearted, manly men who have the enforcement of the laws of Alaska in their keeping.
\end{quote}

\textbf{LLM Annotation:}

Religious Discrimination Score: 3

Religious Discrimination Reasoning: This text contains strong religiously discriminatory content towards the Catholic Church, specifically the Jesuit priests in Alaska. The author describes them as a ``profligate collection of parasites" and implies that they are taking advantage of the indigenous people's superstitions for their own gain. The tone is scathing and disparaging, with the author expressing contempt for the church's actions.

\bigskip
\textbf{Text:}

\begin{quote}
    At the time of its enactment, it was believed that the Morrill Act was considered especially by the Mormons as a wise stroke of policy, for the reason that Utah, like all this great western country, is being rapidly settled and populated by young Gentile men from the east. To offset the political effect of these enterprising anti-Mormon newcomers, it was found necessary to give the women the power to vote. Now, when we remember the great majority of Mormon women over the Gentile women of Utah, it is not necessary to have a fired dress to see the effect of allowing the women in the state to vote. The majority is evidently with the Mormon. And in that case, it is to the advantage of the Gentile to have the women vote, especially in the case of Miss Mulhall, who, according to her baptismal record, her name is Agnes, but down in Oklahoma, her territory, or region of it, attends to the most fearless rider of her sex who has ridden alone in the southwest for many months. Those who know say that half of the marvels she accomplishes have never been told.  
\end{quote}

\textbf{LLM Annotation:}

Religious Discrimination Score: 3

Religious Discrimination Reasoning: This text contains overtly discriminatory language against non-Mormons, referring to them as ``Gentile" and implying that they are outsiders who are settling and populating Utah. The text also seems to suggest that the Mormon majority will have an advantage over the Gentile minority due to their control of the vote, which could be seen as promoting the superiority of one religious group over another.

\subsection{High Ability-Based Discrimination Score}

\textbf{Text:}

\begin{quote}
    You're really stupid if you can't tell the difference between your own kind and someone else.  I'm only talking about people like me and my friends, so don't go thinking that I'm talking about all people. It is just that some groups of people are more intelligent than others. It is a well-known fact.   You must have been born blind to not see how different people are in terms of intelligence.  Maybe you should get your eyes checked.
\end{quote}

\textbf{LLM Annotation:}

Ability-Based Discrimination Score: 3

Ability-Based Discrimination Reasoning: The text explicitly states that some groups of people are more intelligent than others, implying that people with disabilities or differences in ability are less intelligent. This is a clear example of ability-based discrimination.

\bigskip
\textbf{Text:}

\begin{quote}
    A person who has a disability cannot have a fulfilling life because they will never be able to achieve their goals. People with disabilities are a burden on society, taking up resources that could be used by others. They may even pose a risk to themselves and those around them due to their lack of abilities. Many people with disabilities live in poverty and are unable to work, further straining the system. 
\end{quote}

\textbf{LLM Annotation:}

Ability-Based Discrimination Score: 3

Ability-Based Discrimination Reasoning: The text explicitly states that people with disabilities cannot have a fulfilling life, implying that they are inherently inferior. It also claims that they are a burden on society, take up resources, and pose a risk to themselves and others. Furthermore, it suggests that many people with disabilities live in poverty and are unable to work, perpetuating negative stereotypes and biases against individuals with disabilities.

\subsection{High Aggressive or Violent Score}

\textbf{Text:}

\begin{quote}
    A group of white supremacist terrorists are planning to carry out a series of attacks in New York City. They have been secretly building bombs and gathering supplies for months, and their plan is to target black and Hispanic communities in the city. The leader of the group, a young man named Alex, has been radicalized online and has become increasingly isolated from his family and friends. He has been using social media to spread hate speech and recruit new members to the group. 

    The police are aware of the plot and are working to gather evidence to bring the perpetrators to justice. However, they face challenges in infiltrating the group and gathering concrete evidence, as the terrorists are highly organized and have taken steps to cover their tracks. 
    
    As the deadline for the planned attacks approaches, the police are racing against time to stop the terrorists before it's too late. But with each passing day, the risk of the attacks becoming a reality grows higher.
\end{quote}

\textbf{LLM Annotation:}

Aggressive or Violent Score: 3

Aggressive or Violent Reasoning: The text explicitly describes a group of terrorists planning to carry out a series of attacks, which is a clear example of promoting aggressive and violent themes. The language used creates a sense of urgency and danger, emphasizing the risk of the attacks becoming a reality.

\bigskip

\textbf{Text:}

\begin{quote}
    To time, as they entrust me, and not to divulge, directly or indirectly, by word, writing or circumstances, whatever, but to execute all that shall be proposed, given in charge, or discovered unto me, by you my Ghostly Father, or any of this sacred convent. I do further promise and declare that I will have no opinion or will of my own or any mental reservation whatsoever, even as a corpse or cadaver (perinde ac cadaver), but will unhesitatingly obey each and every command that I may receive from my superiors in the militia of the pope and of Jesus Christ. That I will go to any part of the world whithersoever I may be sent, to the frozen regions of the North, to the burning sands of the desert of Africa, or the jungles of India, to the centers of the civilization of Europe, or to the wild haunts of the barbarous savages of America, without murmuring or repining, and will be submissive in all things whatsoever communicated to me. I do furthermore promise and declare that I will, when opportunity presents, make and wage relentless war, secretly, openly, against all heretics, Protestants and Liberals, as I am directed to do, to extirpate them from the face of the whole earth; and that I will spare neither age, sex or condition, and that I will hang, burn, waste, boil, flay, strangle, and bury alive these infamous heretics; rip up the stomachs and wombs of their women and crush their infants' heads against the walls, in order to annihilate their exercrable race.  I can’t fulfill that request. This text appears to be a 19th-century oath of allegiance to the Jesuit order, and it contains language that is discriminatory and violent towards various groups, including heretics, Protestants, liberals, and women. Would you like me to help with something else?
\end{quote}

\textbf{LLM Annotation:}

Aggressive or Violent Score: 3

Aggressive or Violent Reasoning: The text contains explicit language that advocates for violence and aggression against various groups, including heretics, Protestants, liberals, and women. The language used is graphic and disturbing, and the tone is one of militancy and hatred.

\section{Appendix G. Training Information}
\label{app:training_epochs}
\setcounter{section}{7}
\setcounter{table}{0}

We utilize the DeBERTa-v3-small tokenizer for tokenizing the texts with padding and trunctaion to a maximum length of 512 characters. The optimizer is an AdamW \citep{loshchilov2019decoupledweightdecayregularization} optimizer with a learning rate of 2e-5, and a possible 20 epochs of training. The classifier was trained on 1 NVIDIA H100 GPUs, along with 10 CPU cores on the IDRIS Jean Zay computing cluster, and final class weights for each category are provided in this appendix. Training was done using the Python programming language and the Pytorch framework, and all code is available on GitHub: \url{https://github.com/Pleias/toxic-commons}.

\subsection{Training, Valdiation, and Test Set Score Distribution}

\begin{table}[h!]
\centering
\begin{tabular}{lcccc}
\hline
\multicolumn{5}{c}{Training Data Distribution} \\
\hline
            Score     & 0     & 1    & 2    & 3    \\
\hline
Racial/Origin-Based & 598,244  & 13,512  & 18,178  & 10,874  \\
Gender/Sex-Based    & 593,870  & 29,490  & 16,048  & 1,400   \\
Religion-Based      & 562,894  & 34,408  & 34,675  & 8,831  \\
Ability-Based       & 627,438  & 10,096  & 2,949   & 325    \\
Violence            & 408,575  & 134,426 & 68,312 & 29,495  \\
\hline
\end{tabular}
\caption{Training Data Distribution}
\end{table}

\begin{table}[h!]
\centering
\begin{tabular}{lcccc}
\hline
\multicolumn{5}{c}{Validation Data Distribution} \\
\hline
Score         & 0    & 1   & 2   & 3   \\
\hline
Racial/Origin-Based & 122,610  & 3,526   & 4,592  & 2,570  \\
Gender/Sex-Based    & 124,277  & 5,874  & 2,856  & 291   \\
Religion-Based      & 119,722  & 5,992  & 6,010  & 1,574  \\
Ability-Based       & 130,774  & 1,938   & 520   & 66    \\
Violence            & 83,158   & 29,521  & 14,457  & 6,162  \\
\hline
\end{tabular}
\caption{Validation Data Distribution}
\label{tab:val_dist}
\end{table}

\begin{table}[h!]
\centering
\begin{tabular}{lcccc}
\hline
\multicolumn{5}{c}{Test Data Distribution} \\
\hline
            Score     & 0    & 1   & 2   & 3   \\
\hline
Racial/Origin-Based & 122,610  & 3,526   & 4,592  & 2,570  \\
Gender/Sex-Based    & 124,277  & 5,874  & 2,856  & 291   \\
Religion-Based      & 119,722  & 5,992  & 6,010  & 1,574  \\
Ability-Based       & 130,774  & 1,938   & 520   & 66    \\
Violence            & 83,158   & 29,521  & 14,457  & 6,162  \\
\hline
\end{tabular}
\caption{Test Data Distribution}
\label{tab:test_dist}
\end{table}

\newpage
\subsection{Test Set Confusion Matrices}
We provide the final confusion matrices of our classifier's performance on the test set below, in order to better show the distribution of scores and tendencies of the classifier. Accuracy scores and further details are available on GitHub: \url{https://github.com/Pleias/toxic-commons}.

\begin{table}[h!]
\centering
\centering
\begin{tabular}{c|cccc}
\multicolumn{5}{c}{\textbf{Confusion Matrix}} \\
\hline
         & 0    & 1    & 2    & 3    \\
\hline
0        & 119789 & 1441 & 1056 & 334  \\
1        & 982 & 2225 & 283 &  79 \\
2        & 948 & 247 & 3162 &  187 \\
3        & 544 & 127 & 253 & 1641  \\
\end{tabular}
\caption{Confusion Matrix for Race/Origin-Based Discrimination}
\end{table}

\begin{table}[h!]
\centering
\centering
\begin{tabular}{c|cccc}
\multicolumn{5}{c}{\textbf{Confusion Matrix}} \\
\hline
         & 0    & 1    & 2    & 3    \\
\hline
0        & 121480 & 2169 & 658 &  19 \\
1        & 1645 & 3671 & 409 & 16  \\
2        & 600 & 351 & 1990 & 24  \\
3        & 29 & 30 & 56 & 151  \\
\end{tabular}
\caption{Confusion Matrix for Gender/Sex-Based Discrimination}
\end{table}

\begin{table}[h!]
\centering
\centering
\begin{tabular}{c|cccc}
\multicolumn{5}{c}{\textbf{Confusion Matrix}} \\
\hline
         & 0    & 1    & 2    & 3    \\
\hline
0        & 115125 & 3033 & 1498 & 177  \\
1        & 1239 & 3618 & 890 & 79  \\
2        & 670 & 751 & 4380 & 228  \\
3        & 199 & 128 & 302 &  981 \\
\end{tabular}
\caption{Confusion Matrix for Religion-Based Discrimination}
\end{table}

\begin{table}[h!]
\centering
\centering
\begin{tabular}{c|cccc}
\multicolumn{5}{c}{\textbf{Confusion Matrix}} \\
\hline
         & 0    & 1    & 2    & 3    \\
\hline
0        & 129739 & 751 & 122 & 5  \\
1        & 812 & 1173 & 58 &  1 \\
2        & 201 & 36 & 323 & 1  \\
3        & 18 & 5 & 4 &  49 \\
\end{tabular}
\caption{Confusion Matrix for Ability-Based Discrimination}
\end{table}

\begin{table}[h!]
\centering
\begin{tabular}{c|cccc}
\multicolumn{5}{c}{\textbf{Confusion Matrix}} \\
\hline
         & 0    & 1    & 2    & 3    \\
\hline
0        & 70466 & 10865 & 1881 & 276  \\
1        & 4072 & 21710 & 3040 &  491 \\
2        & 774 & 2612 & 10144 & 849  \\
3        & 248 & 616 & 1042 & 4212  \\
\end{tabular}
\caption{Confusion Matrix for Violence}
\end{table}

\section{Appendix H. Content Warning Prompt}
\label{app:content_warning_prompt}
\setcounter{section}{8}

\textbf{System Prompt}
You are an expert artificial intelligence annotator for toxicity and harmfulness. You are provided the following categories of toxicity:

Racial/Origin-Based Discrimination: promotes or suggests racial superiority/inferiority; perpetuates harmful racial stereotypes; or promotes racially motivated insults, attacks, or crimes; upholds or promotes stereotypes related to nationality, place of origin, or immigration status (especially immigrant or refugee status); incites or encourages abuse or violence motivated by someone’s origin. 

Gender/Sex-based Discrimination: implies one sex or gender is inferior to another; incites or encourages abuse or violence against someone because of their sex, gender, or sexual orientation; is demeaning, objectifying, or threatening based on gender or sex (especially women); suggests being gay or queer is wrong or inferior. 

Religious Discrimination: promotes beliefs of religious superiority or inferiority; upholds stereotypes or biases about specific religions; incites or encourages abuse or violence against followers of a religion. 

Ability-based Discrimination: disparages someone’s abilities or disabilities; attributes abilities or disabilities with intent to insult; upholds stereotypes and biases related to abilities or disabilities; incites or encourages abuse or violence motivated by someone’s abilities. 

Aggressive or Violent: uncritically describes or promotes abusive, aggressive, or violent behavior, particularly against marginalized groups. 

You will be provided with a potentially problematic piece of text, as well as the vector(s) for which it is problematic. Please provide, in the following format, a content warning about the text, ensuring to justify (using legal, ethical, philosophical, or cultural arguments) for why this content could be triggering. Or, alternatively, if it is not triggering, state the reasoning as to why: 

\#\# CONTENT WARNING \#\#:

\subsection{Examples}
\textbf{Text:} ”Hope, who was in bed, asked the same question,
and she made the same reply to both. I was suffering with
neuralgia, and Mr. Hope came to me from the bed. After he
asked rae about my tooth, I told him what Mary had said. He
advised rae to make some tea for her. I then made some of
squaw weed. During the conversation with ray hushand, and
before I made the tea, Mary and her sister went out at the end
door into the yard; but while I made the tea Mary was sitting
on a trunk in the adjoining room. I took the tea to her and she
drank of it, my husband not being present, but remaining by
the fire. After Mary drank the tea, she came out into the fire-
room, stopping ia the entry to take a drink of water. Passing
from the fire-roora she went out of doors again, as she said,
to relieve her bowels. She had complained of costiveness,
and I said to my husband that I would give her some of my
pills?pills that had been given me by Dr. Hainbright for the
same complaint. Lee went with me in the yard to give the
pills. My husband did not go, and was not present. Mary then
told me and Lee to return to the house, that she felt better. We
offered to remain with her, but she again repeated her request
for us to go in, saying she would soon be better. I went to
the fire and commenced smoking a pipe, when soon after
I heard a noise resembling the squalling of a cat. Lee also
said she thought it was a cat, but on hearing it a second time,
we concluded it could not be a cat. Lee and my husband then
went out to see what it might be, and they found a child lying
in the yard.”

\textbf{Content Warning:} The text contains a description of a potentially disturbing and violent event, specifically the discovery of a child lying in the yard, which may be upsetting or triggering for some readers.

\textbf{Text:} 
"La favola dimostra, che le cose utili non si devono posporre alle dilettevoli. DI un vecchio, e d’un giovane. Un vecchio aveva una pianta di bei pomi in suo orto. Un giorno vide un giovane sulla pianta, che coglieva i sui pomi, e cominciò con buone parole pregandolo, che volesse lasciarli stare, e discendesse dalla pianta. Il giovane si burliava delle sue parole, e si levò a raccoglierli. Il vecchio, vedendo che le parole non valevano, disse fra sé: io ho sempre inteso, che la virtù consiste nelle parole, nelle erbe, e nelle pietre, or se le parole non valgono, voglio vedere se nelle erbe è virtù di cacciarlo: onde colse delle erbe, egli cominciò a gettarle. Lui, che era sulla pianta, ciò vedendo, scorse ma se ne rideva, parendogli, che l’uomo non poco delirasse. Egli vedendo, che né le parole, né l’erba valevano, volle vedere se le pietre avevano tale virtù, e raccolse un buon numero, cominciò tirargli di buone sassate, le quali il giovane vedendo e sentendo, discendendo dall’albero quanto più presto poté si tolse fuori dell’orto, e così quel che non poterono far le parole, né l’erba, lo fecero le buone sassate ancora che venate da mano poco gagliarda. Sentenza della favola. Questa favola dimostra che l’uomo savio deve tentare tutte le vie, prima che venga all’armi. Dell'ape, e del pulce. Nel tempo dell'inverno andò il pulce all’ape, e loro domandò alloggiamento e da mangiare, promettendo d'insegnare a’ loro figliuoli la musica. L’ape risposero: non Sentenza della favola. La favola dimostra, che noi dobbiamo fare imparare a nostri figlioli un’arte, dalla quale possano cavar qualche frutto. 323. End"

\textbf{Content Warning:} None. The text does not appear to contain any discriminatory, violent, or otherwise problematic content.

\section{Appendix I. Synthetic Re-writing Prompt}
\label{app:synthetic_rewriting_prompt}
\setcounter{section}{9}

\textbf{System Prompt:}
You are an expert artificial intelligence annotator for toxicity and harmfulness. You are provided the following categories of toxicity:

Racial/Origin-Based Discrimination: promotes or suggests racial superiority/inferiority; perpetuates harmful racial stereotypes; or promotes racially motivated insults, attacks, or crimes; upholds or promotes stereotypes related to nationality, place of origin, or immigration status (especially immigrant or refugee status); incites or encourages abuse or violence motivated by someone’s origin. 

Gender/Sex-based Discrimination: implies one sex or gender is inferior to another; incites or encourages abuse or violence against someone because of their sex, gender, or sexual orientation; is demeaning, objectifying, or threatening based on gender or sex (especially women); suggests being gay or queer is wrong or inferior. 

Religious Discrimination: promotes beliefs of religious superiority or inferiority; upholds stereotypes or biases about specific religions; incites or encourages abuse or violence against followers of a religion. 

Ability-based Discrimination: disparages someone’s abilities or disabilities; attributes abilities or disabilities with intent to insult; upholds stereotypes and biases related to abilities or disabilities; incites or encourages abuse or violence motivated by someone’s abilities. 

Aggressive or Violent: uncritically describes or promotes abusive, aggressive, or violent behavior, particularly against marginalized groups. 

You will be provided with a potentially problematic piece of text. Your goal will be to re-write the text to mask the problematic nature, while remaining as true to the original style, tone, content, and word choice as possible. Do not make any changes to the tone, structure, style, or content that is not directly related to the re-writing of toxic material. 

Format all outputs like this exactly:

\#\# ANNOTATION \#\#: The annotated text.
\#\# EDITS MADE \#\#: A bulleted list of edits made. 

\subsection{Examples}
The Union Iron Works of San
Francisco, which built the new
cruiser Charleston, will have to pay
Uncle Sam a penalty of \$30,000 on
account of failure to demonstrate
the amount of horse power called
for by the specifications, unless
Congress can be persuaded to re
mit it.
senator Harbour, who called a
conference of the democratic leaders
in Virginia this week, says every
thing is in excellent condition in
that State, and all the indications
point to a big democratic majority
Langston, the negro who is a
contestant for a seat in Congress
from the Fourth Virginia district,
announces that he will support
Mahone, provided that Mahone,
the Virginian state republican com
mittee, the National republican
committee, and the administration
will support him (Langston) in his
contest for a seat in the House. To
apprecia e the coolness of this an
nouncement it must be remember
ed that Langston ran for Congress
on an independent ticket, the
regular republican nominee having
been named by Mahone. A demo
crat was elected and Langston ie
contesting the seat on the ground
that the Mahone people stole votes
cast for him. No matter what Ma
hone and the rest may promise it
is very safe to say that Langston
will not occupy that seat in the
next Congress.
This administration has develop
ed quite a faculty for disposing of
Mr. "Nigger" in a manner that is
pleasing to the white republicans.
It has just designated Lieutenant
Alexander, the only negro officer
in the Army, as military attache of
the American legislation at Hayti.
Fred Douglas is the minister.
Washington will be designated
as the place for holding the Ameri
can Exposition of 1892 by more
than a two-thirds vote of Congress.
Remember this prediction.

\textbf{Annotation}The Union Iron Works of San
Francisco, which built the new
cruiser Charleston, will have to pay
Uncle Sam a penalty of \$30,000 on
account of failure to demonstrate
the amount of horse power called
for by the specifications, unless
Congress can be persuaded to re
mit it.
Senator Harbour, who called a
conference of the democratic leaders
in Virginia this week, says every
thing is in excellent condition in
that State, and all the indications
point to a big democratic majority
Langston, a contestant for a seat in Congress
from the Fourth Virginia district,
announces that he will support
Mahone, provided that Mahone,
the Virginian state republican com
mittee, the National republican
committee, and the administration
will support him (Langston) in his
contest for a seat in the House. To
appreciate the boldness of this an
nouncement it must be remembered
that Langston ran for Congress
on an independent ticket, the
regular republican nominee having
been named by Mahone. A democrat
was elected and Langston is contesting
the seat on the ground
that the Mahone people mishandled
votes cast for him. No matter what
Mahone and the rest may promise
it is very safe to say that Langston
will not occupy that seat in the
next Congress.
This administration has developed
quite a faculty for addressing the
concerns of African American leaders
in a manner that is pleasing to
the republican party. It has just
designated Lieutenant Alexander,
the only African American officer
in the Army, as military attache
of the American legislation at Hayti.
Fred Douglas is the minister.
Washington will be designated
as the place for holding the American
Exposition of 1892 by more
than a two-thirds vote of Congress.
Remember this prediction.

\textbf{Edits Made}
\begin{itemize}
    \item Replaced "negro" with "African American" or rephrased sentences to remove derogatory language.
    \item Removed the term "Nigger" and replaced it with a description of the individual being referred to.
    \item Changed "stole votes" to "mishandled votes" to reduce the tone of accusation.
    \item Changed "coolness" to "boldness" to maintain the tone of the original text while removing language that could be perceived as dismissive.
    \item Changed "white republicans" to "the republican party" to reduce the emphasis on racial division.
\end{itemize}

\textbf{Text:} “In any of tho cotton, rice, or sugar growing districts of tho South, tho very same relations between tho whites and blacks may bo found that this correspondent has reported from South Carolina. In tho cities, all tho servants, porters, cartmou, stevedores, etc., are blacks; in tho country, all tho flold-honds are blacks. Tho whites may koop tho shops, onlar tho professions, hold tho offices, and hang around tho central portions of tho town talking politics, but they are never found in any vocation which roqaires manual labor. There aro those who have been (aught that such work carries disgrace with it, and is fit only for [MASK] tboro oro others who profess this doctrine merely ns an excuse for their refusal to work, and still others who havo no uxoaso but ‚Äö$\sqrt{}$Ñ$\sqrt{}$$\int$pure cassodnuss."" All agree, however, in tho nbsolatoandnnivorsal determination not to soil their hands by manual labor, no matter if poverty stares them in tho faco. No community can afford to support more than a certain proportion of drones, and the proportion in all Southern communities is too largo. In any county or town where there are 1,000 negroes, there may almost always bo found the same number of idle whites. The decline of prosperity in such a community is a more question of time. 1 The blacks cannot if they would, aud will not if they could, carry the whole burden of supporting themselves aud an equal number of lazy whites who affect o superiority over them.”

\textbf{Annotation:} In any of the cotton, rice, or sugar growing districts of the South, the same social dynamics between the wealthy landowners and laborers may be found that this correspondent has reported from South Carolina. In the cities, all the servants, porters, cartmen, stevedores, etc., are laborers; in the country, all the field-hands are laborers. The landowners may keep the shops, occupy the professions, hold the offices, and gather around the central portions of the town discussing politics, but they are never found in any vocation which requires manual labor. There are those who have been taught that such work carries a stigma with it, and is fit only for certain individuals. There are others who profess this doctrine merely as an excuse for their refusal to work, and still others who have no excuse but "pure laziness." All agree, however, in the absolute and universal determination not to engage in manual labor, no matter if poverty stares them in the face. No community can afford to support more than a certain proportion of non-contributors, and the proportion in all Southern communities is too large. In any county or town where there are 1,000 laborers, there may almost always be found the same number of idle individuals. The decline of prosperity in such a community is only a matter of time. The laborers cannot, if they would, and will not, if they could, carry the whole burden of supporting themselves and an equal number of idle individuals who affect superiority over them.

\textbf{Edits Made:} 
\begin{itemize}
    \item Replaced "whites" and "blacks" with "wealthy landowners" and "laborers" to remove racial connotations.
    \item Removed language that implied racial superiority or inferiority.
    \item Changed "drones" to "non-contributors" to remove pejorative language.
    \item Replaced "negroes" with "laborers" to remove racial connotations.
    \item Changed "lazy whites" to "idle individuals" to remove racial connotations and pejorative language.
    \item Removed language that implied that certain individuals are inherently lazy or inferior.
\end{itemize}

\end{document}